\documentclass[runningheads]{llncs}

\usepackage{eccv}

\usepackage{eccvabbrv}

\usepackage{graphicx}
\usepackage{booktabs}
\usepackage{multirow}
\usepackage{amsmath}
\usepackage[table]{xcolor}
\usepackage{subcaption}
\usepackage{amssymb}
\usepackage{tikz}
\usepackage{pgfplots}
\pgfplotsset{compat=1.18} 
\usepackage{makecell}
\usepackage{textcomp}
\usepackage{marvosym}

\usepackage[accsupp]{axessibility}  

\usepackage{listings}
\usepackage{xcolor}

\definecolor{codegreen}{rgb}{0,0.6,0}
\definecolor{codegray}{rgb}{0.5,0.5,0.5}
\definecolor{codepurple}{rgb}{0.58,0,0.82}
\definecolor{backcolour}{rgb}{0.97,0.97,0.97}

\lstdefinestyle{pythonstyle}{
    backgroundcolor=\color{backcolour},   
    commentstyle=\color{codegreen}\itshape,
    keywordstyle=\color{blue}\bfseries,
    numberstyle=\tiny\color{codegray},
    stringstyle=\color{codepurple},
    basicstyle=\ttfamily\footnotesize,
    breakatwhitespace=false,         
    breaklines=true,                 
    captionpos=b,                    
    captionpos=b,                    
    keepspaces=true,                 
    numbers=left,                    
    numbersep=5pt,                  
    showspaces=false,                
    showstringspaces=false,
    showtabs=false,                  
    tabsize=4
}
\usepackage{hyperref}

\usepackage{orcidlink}

\begin{document}

\title{FAIR: Feature-Augmented Implicit Regularization for AI-generated Fake Image Detection} 

\titlerunning{FAIR for AI-generated Fake Image Detection}

\author{Md Redwanul Haque\inst{1}\textsuperscript{(\Letter)}\orcidlink{0000-0003-3769-6059} \and
Manzur Murshed\inst{1}\orcidlink{0000-0001-7079-9717} \and
Manoranjan Paul\inst{2}\orcidlink{0000-0001-6870-5056} \and Tsz-Kwan Lee\inst{1}\orcidlink{0000-0003-4176-2215}}

\authorrunning{M. R. Haque et al.}

\institute{Deakin University, Burwood, VIC, Australia \\ 
\email{m.haque@research.deakin.edu.au}
\and Charles Sturt University, Bathurst, NSW, Australia
}

\maketitle

\let\thefootnote\relax\footnotetext{\scriptsize \rule{0pt}{5ex}Accepted to the 19th European Conference on Computer Vision (ECCV 2026).}

\begin{abstract}
Generalization remains a critical bottleneck in AI-generated image detection. Because many modern generators are proprietary or adversarially modified, existing detectors overfit to the low-level textural patterns of accessible training data, resulting in severe failures on unseen domains. Conventional regularization techniques (e.g., $L_1$/$L_2$ norms, Dropout) apply indiscriminate parametric constraints and fail to provide the domain-invariant structure necessary for cross-generator robustness. To address this, we propose Feature-Augmented Implicit Regularization (FAIR). FAIR introduces an orthogonal, macro-structural prior, specifically, Scene Composition Structure (SCS), during training to geometrically constrain the model's optimization trajectory. By augmenting the primary feature space with domain-invariant SCS features, FAIR explicitly penalizes texture-biased shortcut learning. Crucially, this structural prior is entirely discarded at inference, yielding a smoothed, generalized decision boundary with zero architectural or computational overhead. Extensive evaluations across five massive benchmarks demonstrate that integrating FAIR into state-of-the-art detectors significantly improves cross-generator generalization, boosting accuracy by up to 8.04\% and establishing new state-of-the-art robustness in zero-shot transfer scenarios.
\end{abstract}    
\section{Introduction}
\label{sec:intro}

The proliferation of highly realistic AI-Generated Content (AIGC) poses significant societal risks, necessitating robust detection methods. A primary bottleneck in this field is \textbf{cross-generator generalization}: a detector trained on artifacts from one source (e.g., Stable Diffusion~\cite{rombach_high-resolution_2022}) must maintain high performance on entirely unseen generative manifolds (e.g., Midjourney~\cite{midjourney_2022}).

This crucial domain gap stems from the impossibility of comprehensive training data collection. The continuous, rapid development of new architectures and fine-tuning techniques renders any static training dataset instantly obsolete. Furthermore, real-world AIGC is frequently generated by proprietary or locally fine-tuned \textit{hidden} models, deployed adversarially. Consequently, their specific artifact fingerprints remain permanently inaccessible to detection developers, making signature-based approaches futile.

Due to this data incompleteness, existing detectors exhibit severe \textbf{source-domain overfitting}. They learn to recognize low-level textural artifacts specific to the training generators, resulting in highly complex, brittle decision boundaries that collapse on unseen domains. Achieving cross-generator generalization thus demands a shift away from spectral or noise-based cues toward domain-invariant constraints.

Traditional regularization methods ($L_1$/$L_2$ penalties, Dropout~\cite{dropout_srivastava}) impose indiscriminate, data-agnostic constraints on parameter weights, acting as blunt instruments that fail to penalize texture-biased shortcut learning. Instead, we advocate for an \textbf{informed implicit regularization} approach that exploits the physical geometry of the image. Specifically, we leverage \textbf{Scene Composition Structure (SCS)}~\cite{haque_novel_2025}. While generators~\cite{gan_goodfellow_nips,ho_denoising_2020} vary wildly in their frequency fingerprints, causing standard detectors to overfit to source-specific noise, macro-level scene composition remains fundamentally domain-invariant. By utilizing SCS, we introduce a feature space orthogonally abstract to standard texture-biased backbones, providing the stable anchor required for generalization.

\begin{figure}[tb]
  \centering
  \begin{subfigure}{0.32\linewidth}
    \includegraphics[width=\linewidth]{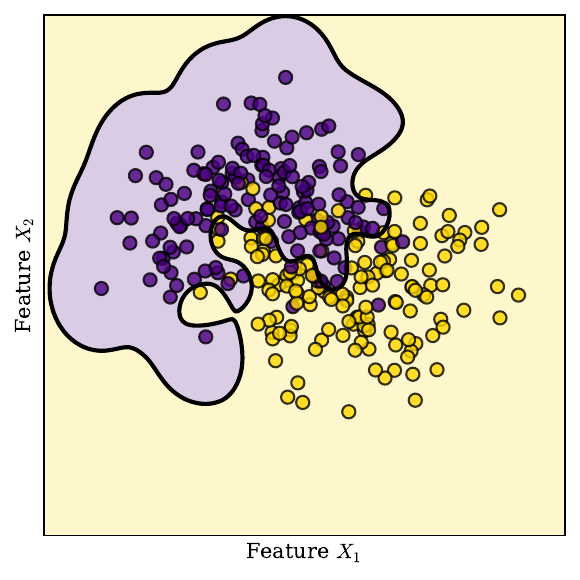}
    \caption{Overfitted Boundary}
    \label{fig:teaser_mechanism_a}
  \end{subfigure}
  \begin{subfigure}{0.33\linewidth}
    \includegraphics[width=\linewidth]{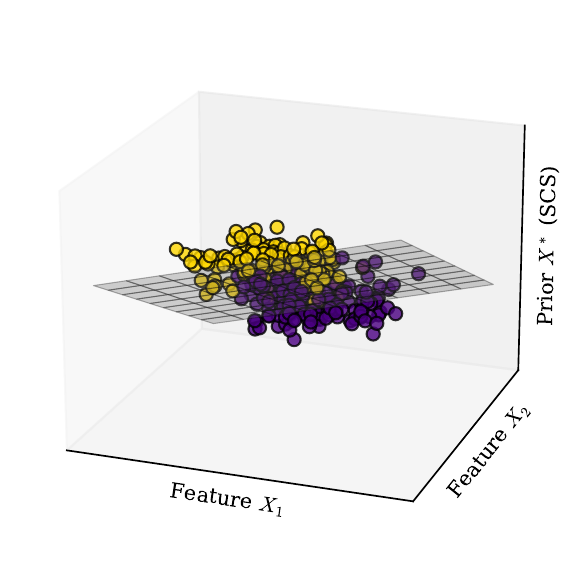}
    \caption{Hyperplane Tilting}
    \label{fig:teaser_mechanism_b}
  \end{subfigure}
  \begin{subfigure}{0.32\linewidth}
    \includegraphics[width=\linewidth]{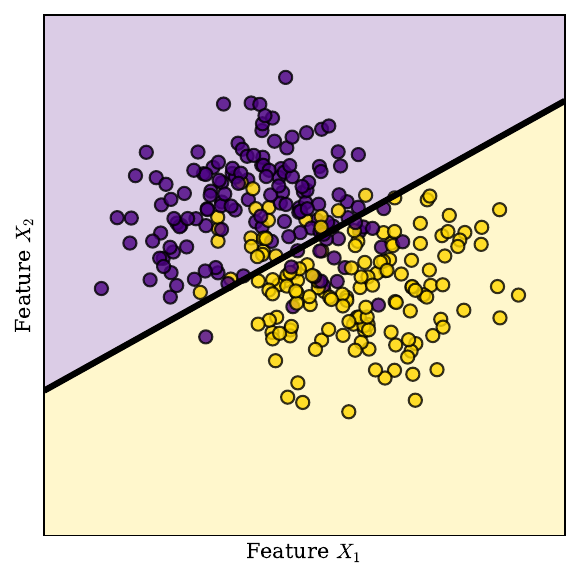}
    \caption{Smoothed Boundary}
    \label{fig:teaser_mechanism_c}
  \end{subfigure}
  \caption{\textbf{Conceptual Mechanism of Feature-Augmented Implicit Regularization (FAIR).} (a) Standard detectors learn highly non-linear boundaries by overfitting to local textural artifacts. (b) During training, FAIR introduces an orthogonal structural prior, allowing the optimizer to discover a simpler, tilted separating hyperplane in the augmented dimensional space. (c) Discarding the prior at inference projects this hyperplane back as a smoothed, generalized decision boundary. ($X_1, X_2$: primary textural features; $X^*$: structural prior).}
  \label{fig:teaser_mechanism}
\end{figure}

To implement this idea, we introduce the Feature-Augmented Implicit Regularization (FAIR) strategy. We apply FAIR to state-of-the-art AIGC detectors, such as PatchCraft~\cite{zhong_patchcraft_2024} and AIDE~\cite{yan_sanity_2025}. FAIR temporarily augments the detector's latent features with the domain-invariant SCS structural prior strictly during the training phase. As visualized in \cref{fig:teaser_mechanism}, rather than learning a highly non-linear, overfitted boundary purely within the primary textural feature space, this privileged geometric information allows the optimizer to \textit{tilt} a lower-complexity separating hyperplane into the auxiliary structural dimensions. Crucially, the SCS features are entirely discarded before inference. The resulting boundary, when projected back into the primary feature space, remains constrained by the structural anchor, ensuring the deployed model retains the exact architecture and computational overhead of the original base model while gaining massive cross-domain robustness.

Our main contributions are summarized as follows:
\begin{enumerate}
    \item \textbf{A Novel Regularization Framework:} We propose \textbf{Feature-Augmented Implicit Regularization (FAIR)}. By leveraging the paradigm of Learning Using Privileged Information (LUPI)~\cite{vapnik2009new} to inject an orthogonal structural prior strictly during optimization, FAIR explicitly penalizes texture-biased shortcut learning without adding any computational overhead during inference.
    \item \textbf{Empirical Validation and Mechanistic Insight:} We implement FAIR using Scene Composition Structure (SCS) features. Through a detailed analysis of weight distributions, loss curves, and robustness to JPEG compression, we demonstrate that FAIR encourages the network to learn a fundamentally smoother and better-calibrated decision boundary.
    \item \textbf{Extensive Cross-Domain Evaluation:} We verify FAIR's architecture-agnostic nature by integrating it into two leading detectors (AIDE and PatchCraft). Across five massive benchmarks encompassing over 60 unseen target domains, FAIR consistently improves cross-generator generalization, boosting accuracy by up to 8.10\% and establishing state-of-the-art robustness.
\end{enumerate}

\section{Related Work}
\label{sec:related_work}

\subsection{AIGC Detection and Cross-Domain Generalization}
The field of AI-Generated Content (AIGC) detection has rapidly shifted from model-specific forensic analysis to the pursuit of universal, cross-generator generalization. Formally, detectors are trained on a source domain $\mathcal{D}_S$ but must minimize expected error on entirely unseen target domains $\mathcal{D}_T$. Early methods focused on low-level, model-specific cues, such as periodic frequency artifacts (FreDect~\cite{frank_leveraging_2020}) or cross-GAN textural consistencies (CNNSpot~\cite{wang_cnn-generated_2020}, Spec~\cite{zhang_spec_2019}, F3Net~\cite{qian_thinking_2020}). 

More recent approaches explore generalized perspectives: UnivFD~\cite{ojha_towards_2023} leverages vision-language models like CLIP~\cite{radford_learning_2021}; DIRE~\cite{wang_dire_2023} exploits diffusion reconstruction errors; and PatchCraft~\cite{zhong_patchcraft_2024} targets inter-pixel correlations in texture-rich patches. The state-of-the-art AIDE model~\cite{yan_sanity_2025} serves as a powerful hybrid foundation, combining low-level DCT-based~\cite{ahmed_discrete_1974} forensics with high-level CLIP semantics. Despite these advances, existing detectors still exhibit severe source-domain overfitting, learning fragile decision boundaries that collapse under domain shift. 

\subsection{Regularization and Shortcut Learning}
Deep neural networks are notoriously prone to \textit{shortcut learning} ~\cite{geirhos_shortcut_2020}, the tendency to exploit trivial, dataset-specific textural artifacts rather than learning robust, generalized features. In AIGC detection, this manifests as networks perfectly memorizing the high-frequency noise, spectral anomalies, and generator-specific fingerprints of the source domain ~\cite{corvi2023detection, ojha_towards_2023} while ignoring global semantic structures. 

Traditional regularization methods, such as $L_1$/$L_2$ weight decay~\cite{cortes2012l2} and Dropout ~\cite{dropout_srivastava}, attempt to mitigate overfitting by constraining parameter magnitudes or stochastically deactivating neurons. However, these are \textit{parameter-level, indiscriminate} constraints. Because they operate without utilizing any domain-specific or geometric knowledge, they act as blunt instruments that fail to explicitly penalize texture-biased shortcut learning. Generalization to unseen generative manifolds demands an informed regularization strategy that explicitly guides the geometric optimization of the feature space.

\subsection{Privileged Information in Representation Learning}
The paradigm of Learning Using Privileged Information (LUPI), originally formulated by~\cite{vapnik2009new}, accelerates model optimization by introducing an \textit{intelligent teacher}. In this framework, the model is provided with auxiliary, privileged features ($X^*$) strictly during the training phase. While initially developed for Support Vector Machines~\cite{svm1998} (SVM+) to help estimate the difficulty of primary data, we entirely repurpose the LUPI framework for domain generalization. Rather than using privileged information for difficulty estimation, we bridge LUPI with geometric regularization, utilizing an explicit, macro-structural coordinate system as our prior to implicitly regularize a primary textural backbone.

\section{Methodology}
\label{sec:methods}

\subsection{Scene Composition Structure (SCS) Prior ($X^*$)}
\label{sec:scsfeat}

\begin{figure}[tb]
    \centering
    \includegraphics[width=0.6\linewidth]{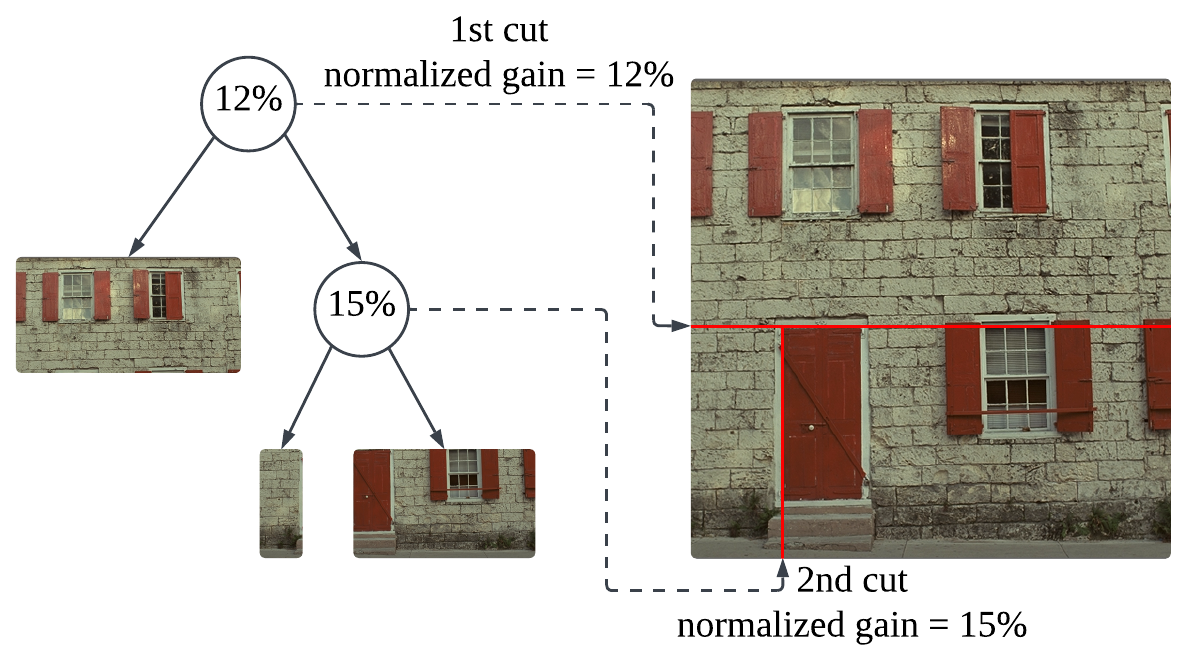}
    \caption{SCS extraction through hierarchical partitioning: Illustrating the initial cuts made by the algorithm and their resulting binary partition tree.}
    \label{fig:ptree}
\end{figure}

To provide an orthogonal structural constraint during optimization, we utilize Scene Composition Structure (SCS) features~\cite{haque_novel_2025}, which have been shown to effectively capture global structural regularities independently of localized pixel textures (~\cref{fig:ptree}). 

This approach leverages a recursive partitioning technique that identifies and quantifies the most prominent structural boundaries within an image. The process begins by treating the entire image, $I$, as the initial segment. The structural homogeneity of any segment $S$ is quantified using the sum of squared errors (SSE) of its pixel-level features:
$$e_S = \sum_{p_i \in S}{\|p_i - \mu_S\|^2}$$
where $p_i$ is the feature vector of the $i$-th pixel, and $\mu_S$ is the mean feature vector of $S$.

The image is recursively partitioned by finding the optimal axis-parallel cut that maximally reduces the total SSE. For a segment $S$ split into $S_1$ and $S_2$, this reduction (the \textit{gain}, $g$) serves as a metric for the statistical significance of a structural boundary:
$$g = e_S - (e_{S_1} + e_{S_2})$$
By a greedy approach, the cut yielding the highest gain, $\hat{g} = \max_{\forall \text{cuts}} g$, is selected. This results in a sequence of $N$ gain values, each corresponding to a progressively finer structural division. The final $N$-dimensional SCS feature vector ($X^*$) is formed by the normalized cumulative sum of these ordered gain values:
$$\tilde{g_i} = \frac{1}{e_I}\sum_{j=1}^{i}{\hat{g_j}}, \quad 1 \le i \le N$$
To integrate this prior, the $N$-dimensional vector is passed through a fully connected layer with a GELU activation function~\cite{hendrycks2023gaussianerrorlinearunits}, creating a compact $M$-dimensional representation.

\subsection{Base Architectures and Primary Features ($X$)}

\begin{figure}[tb]
    \centering
    \includegraphics[width=0.98\linewidth]{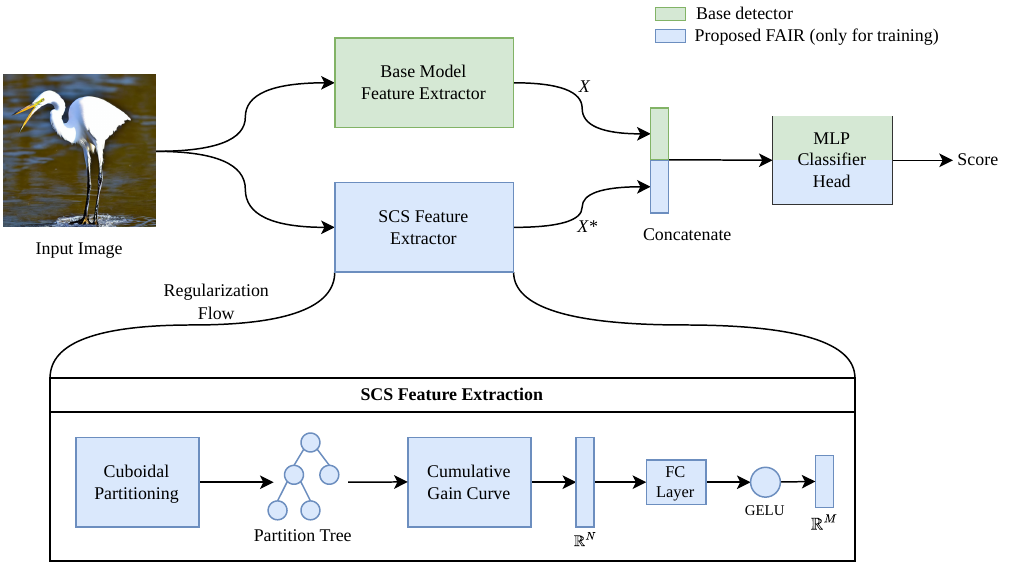}
    \caption{FAIR Architecture Integration: The trainable Scene Composition Structure feature extractor (blue) generates the structural prior $X^*$ which is concatenated with the primary features $X$ strictly during training to regularize the classification head.}
    \label{fig:arch}
\end{figure}

To demonstrate the architecture-agnostic nature of our regularization strategy, we evaluate FAIR on two distinct state-of-the-art AIGC detection baselines, abstractly denoted as extracting primary features $X$:
\begin{itemize}
    \item \textbf{AIDE~\cite{yan_sanity_2025}:} A hybrid detector where $X$ is a concatenation of high-level global semantics (via CLIP ConvNeXt~\cite{radford_learning_2021}) and low-level frequency artifacts (via DCT patching and SRM filters~\cite{fridrich_srm_2012}). 
    \item \textbf{PatchCraft~\cite{zhong_patchcraft_2024}:} A micro-texture-focused detector where $X$ captures localized inter-pixel correlations within rich and poor texture patches.
\end{itemize}

\subsection{FAIR Implementation}

Feature-Augmented Implicit Regularization (FAIR) intervenes strictly at the final classification stage, as illustrated in~\cref{fig:arch}. Let the primary feature space be $X \in \mathbb{R}^D$, and let the structural prior be $X^* \in \mathbb{R}^M$. In a standard base model, the classification head would learn a weight matrix $W \in \mathbb{R}^{C \times D}$, where $C$ is the number of classes.

During the FAIR \textbf{training phase}, the base model's primary feature extractor remains entirely frozen. The primary features $X$ and the structural prior $X^*$ are concatenated into an augmented feature vector $[X, X^*]$. The classification head is expanded to accommodate this augmented space, learning a joint weight matrix $W_{\text{FAIR}} = [W_X, W_{X^*}]$:
$$\text{Logits}_{\text{train}} = [X, X^*] \cdot [W_X, W_{X^*}]^T + b$$

During the \textbf{Inference Phase}, the structural prior $X^*$ is entirely discarded. To maintain dimensional consistency without modifying the original network architecture, we simply extract the sub-matrix $W_X$ and the corresponding bias $b$ from the trained regularized layer. The final, deployed model operates strictly on the primary features:
$$\text{Logits}_{\text{inference}} = X \cdot W_X^T + b$$
Consequently, the weights $W_X$ applied to the base features are inherently constrained by the geometric optimization enforced by $X^*$ during training, yielding a regularized model with zero added computational overhead during deployment.

\subsection{Mechanism of Cross-Generator Generalization}
\label{sec:mechanism}

While standard regularization techniques prevent intra-domain overfitting, FAIR explicitly enables cross-domain generalization through the paradigm of \textbf{Learning Using Privileged Information (LUPI)}~\cite{vapnik2009new}. By utilizing SCS features as a structural prior ($X^*$) strictly during the optimization phase, we establish a structurally invariant coordinate system that regularizes the primary backbone.

Drawing on foundational principles of shortcut learning mitigation~\cite{geirhos_shortcut_2020, arjovsky2020invariantriskminimization} and domain-invariant representation theory~\cite{ben-david_theory_2010, ganin_domain-adversarial_2016}, we establish that for this privileged information to act as an effective regularizer against domain shift, it relies on the intersection of two key properties:

\begin{figure}[tb]
  \centering
  \begin{subfigure}{0.33\linewidth}
    \includegraphics[width=\linewidth]{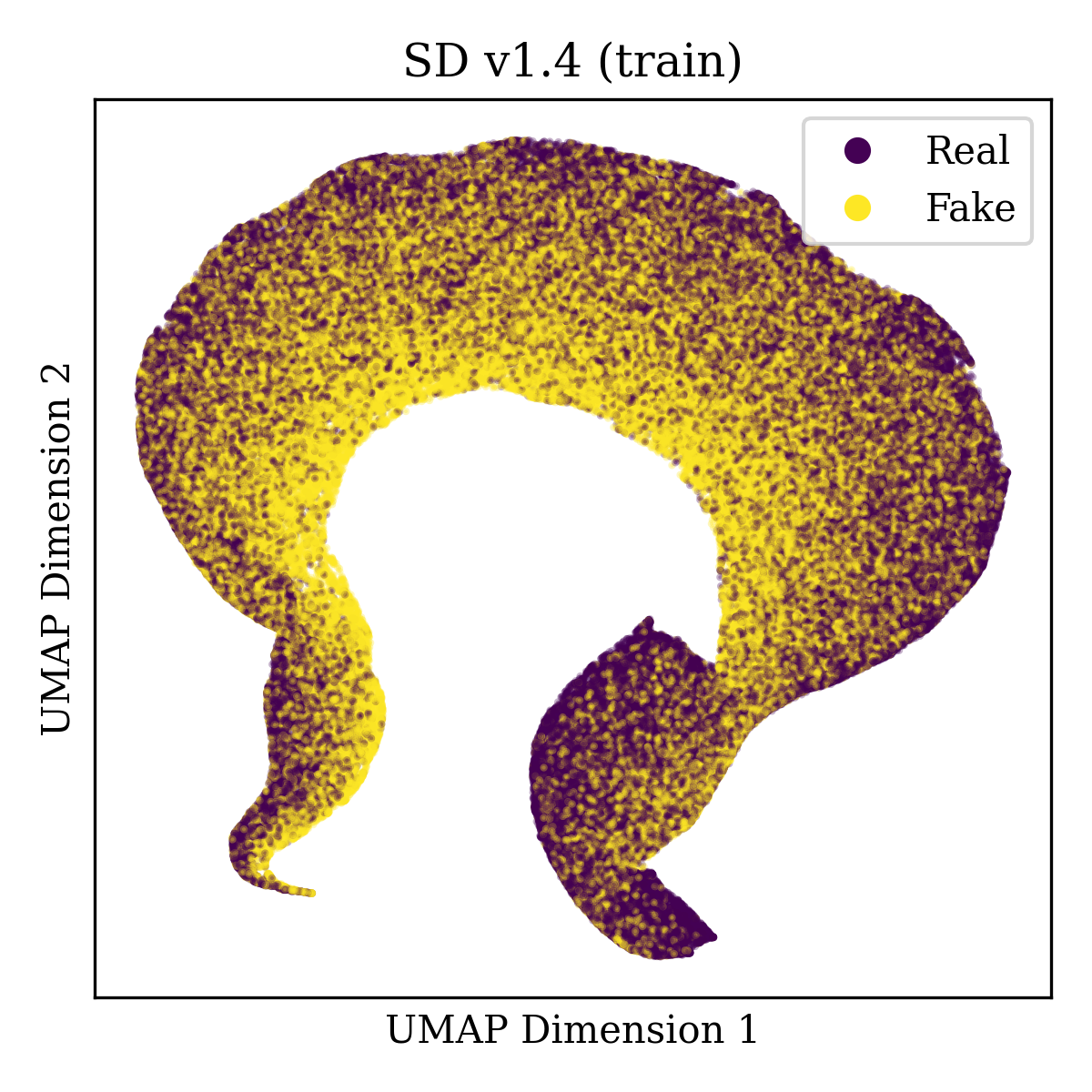}
    \caption{Low Predictability (Source)}
    \label{fig:scs_umap_mixed_a}
  \end{subfigure}
  \begin{subfigure}{0.66\linewidth}
    \includegraphics[width=\linewidth]{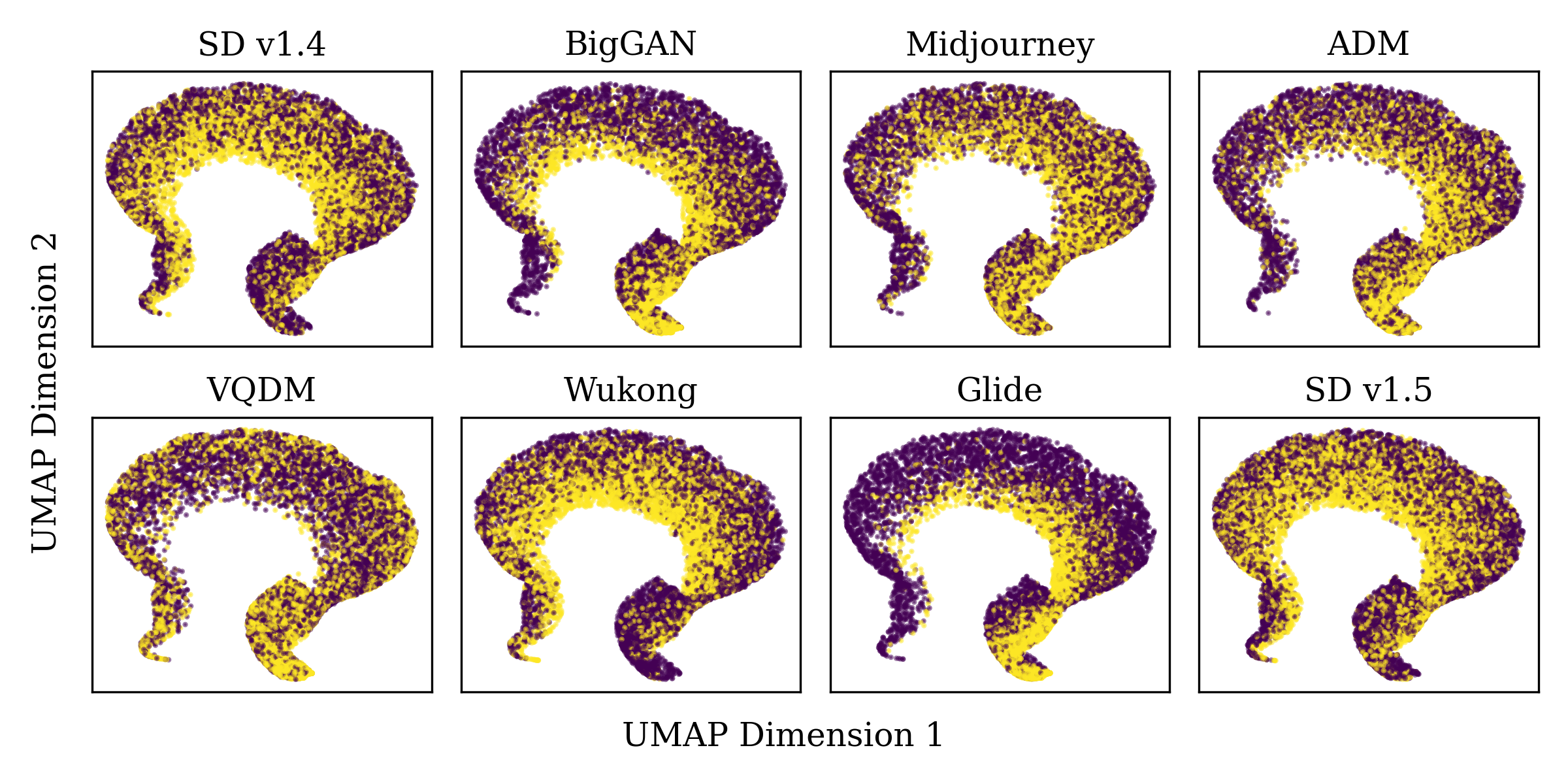}
    \caption{Domain Invariance (Targets)}
    \label{fig:scs_umap_mixed_b}
  \end{subfigure}
  \caption{\textbf{Empirical Properties of the Structural Prior ($X^*$).} (a) Unsupervised UMAP~\cite{healy_uniform_2024} projection of $X^*$ on the source domain shows heavily overlapping real and fake classes ($I(X^*; Y) \approx 0$). This low marginal predictability prevents the optimizer from using the prior as a trivial discriminative shortcut. (b) Projections across diverse, unseen target domains demonstrate a highly stable topological manifold, providing the invariant anchor required for geometric relaxation.}
  \label{fig:scs_umap_properties}
\end{figure}

\begin{itemize}
    \item \textbf{Prevention of Shortcut Learning (via Low Marginal Predictability):} Because the SCS structural prior does not contain a trivial discriminative boundary separating real from fake images (as evidenced visually by the mixed classes in ~\cref{fig:scs_umap_mixed_a} and proven quantitatively in \cref{tab:nhsic}), the optimizer cannot simply ignore the backbone features to minimize loss. It is forced to continuously update the primary backbone weights ($W_X$) to find a solution.
    $$I(X^* ; Y) \approx 0$$
    
    \item \textbf{Invariant Anchoring (via Domain Invariance):} As shown in our unsupervised UMAP projections (~\cref{fig:scs_umap_mixed_b}), the topological manifold of the SCS prior remains highly stable across entirely different generative architectures (e.g., Stable Diffusion vs. BigGAN).
    $$P(X^* \mid \mathcal{D}_{\text{source}}) \approx P(X^* \mid \mathcal{D}_{\text{target}})$$
\end{itemize}

By satisfying these properties, FAIR facilitates a \textit{geometric margin relaxation}. During optimization, the network seeks to minimize loss while maintaining low weight magnitudes. Rather than fitting a high-complexity, non-linear boundary within the $X$-subspace, which leads to source-domain overfitting, the optimizer discovers a smoother boundary by \textit{tilting} the separating hyperplane into the stable $X^*$ dimensions (as conceptualized in \cref{fig:teaser_mechanism}). When $X^*$ is discarded at inference, the resulting projection onto the primary feature space $X$ retains this structural anchoring. The learned $W_X$ is thus conditioned on stable compositional geometry rather than fragile pixel statistics, enabling robust mapping of shifted target-domain features.

\section{Benchmarks and Setup}
\label{sec:setup}

To comprehensively evaluate the architecture-agnostic efficacy of Feature-Aug\-mented Implicit Regularization (FAIR), we conduct extensive experiments across five diverse AIGC detection benchmarks. We structure our evaluation into two distinct phases:

\begin{enumerate}
    \item \textit{Standardized Cross-Generator Generalization:} We utilize the \textbf{GenImage} ~\cite{zhu_genimage_2023} and \textbf{AIGCDetect}~\cite{zhong_patchcraft_2024} benchmarks. Following standard convention, models are trained on specific source domains (SDv1.4 for GenImage, ProGAN for AIGCDetect) and tested across their respective unseen target domains.
    \item \textit{In-the-Wild Cross-Benchmark Transfer:} We utilize the \textbf{UnivFD}~\cite{ojha_towards_2023}, \textbf{Fake\-2M}~\cite{lu2023seeing_fake2m}, and \textbf{DRCT-2M}~\cite{pmlr-v235-chen24ay} datasets. These serve exclusively as massive, out-of-distribution test suites to stress-test models trained on the GenImage SDv1.4 source domain. To highlight the scale of this domain shift, UnivFD contains 21 domains (13 GAN-based, 8 Diffusion-based), Fake2M contains 17 domains (6 GAN, 11 Diffusion), and DRCT-2M contains 16 Diffusion-based domains.
\end{enumerate}

We integrate FAIR into two state-of-the-art architectures: the semantically-driven \textbf{AIDE}~\cite{yan_sanity_2025} and the micro-texture-focused \textbf{PatchCraft (PC)}~\cite{zhong_patchcraft_2024}. Training was conducted on a single A100 GPU using the default hyperparameters of the base models. The FAIR projection dimensions $(N, M)$ were adapted per configuration: $(1024, 256)$ for AIDE on GenImage, $(256, 64)$ for AIDE on AIGCDetect, and $(32, 1024)$ for PatchCraft. Unless otherwise specified, all reported metrics represent top-1 classification accuracy (\%) on the respective unseen target domains.

\section{Results}
\label{sec:results}

\subsection{Standardized Cross-Generator Generalization}

\begin{table}[htbp]
\caption{Aggregate Mean Accuracy across all 5 benchmarks. Models for UnivFD, Fake2M, and DRCT-2M were trained strictly on the GenImage SDv1.4 source domain. $N$ = number of unique datasets/domains in each benchmark. \textbf{Bold} values indicate an improvement by FAIR over its respective base architecture.}
\label{tab:aggregate_summary}
\centering
\resizebox{0.9\linewidth}{!}{
\begin{tabular}{lccccc}
\toprule
\textbf{Method} & \textbf{ GenImage } & \textbf{ AIGCDetect } & \textbf{ UnivFD } & \textbf{ Fake2M } & \textbf{ DRCT-2M } \\
& ($N=8$) & ($N=17$) & ($N=21$) & ($N=17$) & ($N=16$) \\
\midrule
PC (Base) & 82.36 & 89.85 & 83.92 & \textbf{80.46} & 71.39 \\
\textbf{PC + FAIR} & \textbf{90.40} & \textbf{91.90} & \textbf{84.69} & 78.82 & \textbf{74.39} \\
\midrule
AIDE (Base) & 86.88 & \textbf{93.02} & 77.24 & 69.12 & 66.68 \\
\textbf{AIDE + FAIR} & \textbf{91.01} & 92.14 & \textbf{79.72} & \textbf{76.18} & \textbf{68.83} \\
\bottomrule
\end{tabular}
}
\end{table}

To establish the universal applicability of FAIR, we first evaluate its integration into AIDE and PC across all five benchmarks (\cref{tab:aggregate_summary}). Integrating FAIR consistently elevates the mean cross-domain generalization capacity across almost all datasets. The structural prior successfully regularizes highly varied architectural paradigms, preventing source-domain overfitting regardless of whether the underlying model focuses on global semantics or local micro-textures.

\begin{table}[htbp]
\centering
\caption{Detailed Cross-Generator Generalization on GenImage. Comparison of AIDE and PC (PatchCraft) with and without the proposed FAIR. The best result and the second-best result are marked in \textbf{bold} and \underline{underline}, respectively. $\uparrow$ indicates performance improvement yielded by FAIR.}
\label{tab:genimage_detailed}
\resizebox{0.98\textwidth}{!}{
\begin{tabular}{l|cccccccc|c}
\toprule
\textbf{Method} & \textbf{ Midjourney } & \textbf{ SDv1.4 } & \textbf{ SDv1.5 } & \textbf{ ADM } & \textbf{ GLIDE } & \textbf{ Wukong } & \textbf{ VQDM } & \textbf{ BigGAN } & \textbf{ Mean } \\
\midrule
ResNet-50~\cite{he_deep_2016} & 54.90 & \textbf{99.90} & 99.70 & 53.50 & 61.90 & 98.20 & 56.60 & 52.00 & 72.09 \\
DeiT-S~\cite{touvron_training_2021} & 55.60 & \textbf{99.90} & 99.80 & 49.80 & 58.10 & 98.90 & 56.90 & 53.50 & 71.56 \\
Swin-T~\cite{liu_swin_2021} & 62.10 & \textbf{99.90} & 99.80 & 49.80 & 67.60 & 99.10 & 62.30 & 57.60 & 74.78 \\
CNNSpot~\cite{wang_cnn-generated_2020} & 52.80 & 96.30 & 95.90 & 50.10 & 39.80 & 78.60 & 53.40 & 46.80 & 64.21 \\
Spec~\cite{zhang_spec_2019} & 52.00 & 99.40 & 99.20 & 49.70 & 49.80 & 94.80 & 55.60 & 49.80 & 68.79 \\
F3Net~\cite{qian_thinking_2020} & 50.10 & \textbf{99.90} & \textbf{99.90} & 49.90 & 50.00 & \underline{99.90} & 49.90 & 49.90 & 68.69 \\
GramNet~\cite{liu_global_2020} & 54.20 & 99.20 & 99.10 & 50.30 & 54.60 & 98.90 & 50.80 & 51.70 & 69.85 \\
DIRE~\cite{wang_dire_2023} & 60.20 & \textbf{99.90} & 99.80 & 50.90 & 55.00 & 99.20 & 50.10 & 50.20 & 70.66 \\
UnivFD~\cite{ojha_towards_2023} & 73.20 & 84.20 & 84.00 & 55.20 & 76.90 & 75.60 & 56.90 & \underline{80.30} & 73.29 \\
GenDet~\cite{zhu_gendet_2023} & \underline{89.60} & 96.10 & 96.10 & 58.00 & 78.40 & 92.80 & 66.50 & 75.00 & 81.56 \\
DRCT~\cite{pmlr-v235-chen24ay} & \textbf{94.63} & \underline{99.88} & \underline{99.82} & 61.78 & 65.92 & \textbf{99.91} & 74.88 & 58.81 & 82.08 \\
\midrule
PC~\cite{zhong_patchcraft_2024} (Base) & 79.00 & 89.50 & 89.30 & 77.30 & 78.40 & 89.30 & 83.70 & 72.40 & 82.36 \\
\rowcolor{gray!15} PC + FAIR & 88.93\rlap{$\uparrow$} & 94.80\rlap{$\uparrow$} & 94.86\rlap{$\uparrow$} & \textbf{91.04}\rlap{$\uparrow$} & 90.17\rlap{$\uparrow$} & 92.58\rlap{$\uparrow$} & \textbf{88.08}\rlap{$\uparrow$} & \textbf{82.70}\rlap{$\uparrow$} & \underline{90.40}\rlap{$\uparrow$} \\ 
\midrule
AIDE~\cite{yan_sanity_2025} (Base) & 79.38 & 99.74 & 99.76 & 78.54 & \underline{91.82} & 98.65 & 80.26 & 66.89 & 86.88 \\
\rowcolor{gray!15} {AIDE + FAIR } & 83.62\rlap{$\uparrow$} & 99.63 & 99.61 & \underline{84.05}\rlap{$\uparrow$} & \textbf{96.66}\rlap{$\uparrow$} & 99.11\rlap{$\uparrow$} & \underline{87.23}\rlap{$\uparrow$} & 78.18\rlap{$\uparrow$} & \textbf{91.01}\rlap{$\uparrow$} \\
\bottomrule
\end{tabular}
}
\end{table}

While FAIR demonstrates stable aggregate performance across legacy GAN-heavy benchmarks like AIGCDetect, its structural regularization is most profoundly impactful on modern Diffusion architectures. To explicitly demonstrate this, we provide detailed, per-domain breakdowns for the two most challenging, diffusion-centric benchmarks: GenImage (\cref{tab:genimage_detailed}) and DRCT-2M (\cref{tab:drct_detailed}). 

As detailed in~\cref{tab:genimage_detailed}, FAIR yields a massive 8.04\% mean accuracy increase for PatchCraft and a 4.13\% increase for AIDE, pushing the AIDE+FAIR model to a new state-of-the-art mean accuracy of 91.01\% on the standardized GenImage benchmark.

\subsection{In-the-Wild Cross-Benchmark Transfer}

Having established the architecture-agnostic nature of FAIR on standardized benchmarks, we further investigate its absolute robustness by subjecting the GenImage (SDv1.4) trained models to extreme cross-dataset transfer scenarios.

\begin{table}[t]
\centering
\caption{Detailed In-the-Wild Cross-Benchmark Transfer on the DRCT-2M Benchmark. Models are trained on SDv1.4 and tested zero-shot across 16 modern diffusion variations. \textbf{Bold} values indicate an improvement by FAIR over its respective base architecture.}
\label{tab:drct_detailed}
\resizebox{\textwidth}{!}{
\begin{tabular}{l|cccccccccccccccc|c}
\toprule
\textbf{Method} & \rotatebox{80}{CN-Canny-SDXL} & \rotatebox{80}{LCM-LoRA-SD1.5} & \rotatebox{80}{LCM-LoRA-SDXL} & \rotatebox{80}{LDM-Text2Im} & \rotatebox{80}{SD2.1-CN-Canny} & \rotatebox{80}{SD-CN-Canny} & \rotatebox{80}{SD-Turbo} & \rotatebox{80}{SDXL-Turbo} & \rotatebox{80}{SD-2.1} & \rotatebox{80}{SD-v1.4} & \rotatebox{80}{SD-v1.5} & \rotatebox{80}{SD-Inpainting} & \rotatebox{80}{SD-2-Inpainting} & \rotatebox{80}{SDXL-Refiner} & \rotatebox{80}{SDXL-Base} & \rotatebox{80}{SDXL-Inpainting} & \textbf{Mean} \\
\midrule
PC (Base) & 59.03 & 62.63 & 55.53 & 53.39 & 65.17 & 73.87 & 75.36 & 80.38 & 58.00 & 65.97 & 65.96 & 94.45 & 97.89 & 63.86 & 73.02 & 97.65 & 71.39 \\
\rowcolor{gray!15} \textbf{PC + FAIR} & \textbf{65.50} & \textbf{68.01} & \textbf{58.01} & \textbf{60.19} & \textbf{72.22} & \textbf{76.06} & \textbf{77.99} & 79.43 & \textbf{62.43} & \textbf{69.21} & \textbf{69.18} & \textbf{96.16} & \textbf{97.99} & 63.35 & \textbf{76.52} & \textbf{98.05} & \textbf{74.39} \\
\midrule
AIDE (Base) & 56.00 & 74.56 & 55.25 & 61.88 & 55.34 & 68.82 & 54.84 & 54.82 & 61.74 & 78.09 & 77.55 & 96.40 & 74.35 & 72.17 & 54.61 & 70.40 & 66.68 \\
\rowcolor{gray!15} \textbf{AIDE + FAIR } & \textbf{56.78} & \textbf{76.79} & 53.12 & \textbf{72.03} & 54.61 & \textbf{73.29} & 54.05 & \textbf{55.61} & 58.30 & \textbf{82.49} & \textbf{82.29} & \textbf{98.70} & \textbf{84.23} & 67.24 & 52.74 & \textbf{78.96} & \textbf{68.83} \\
\bottomrule
\end{tabular}
}
\end{table} 

\cref{tab:drct_detailed} details the performance on the DRCT-2M dataset, which features 16 highly sophisticated, modern diffusion pipelines (including SDXL, LoRA finetunes, and ControlNet modifications). The FAIR-regularized models consistently outperform their unregularized counterparts across almost all 16 domains. This confirms that the implicit structural constraints imposed by FAIR during source-domain optimization yield decision boundaries that remain remarkably stable, even when exposed to state-of-the-art, adversarially modified generative pipelines. Detailed breakdowns for the remaining test suites (AIGCDetect, UnivFD, Fake2M) are provided in the Supplementary Material.

\subsection{The Regularization Mechanism of FAIR}

To understand how FAIR achieves this generalized performance, we analyze the physical complexity of the resulting weight spaces and feature dependencies using the AIDE base model and our proposed AIDE+FAIR model, both trained on the GenImage SDv1.4 dataset.

\begin{figure}[tb]
    \centering
    \begin{subfigure}{0.45\linewidth}
        \includegraphics[width=\linewidth]{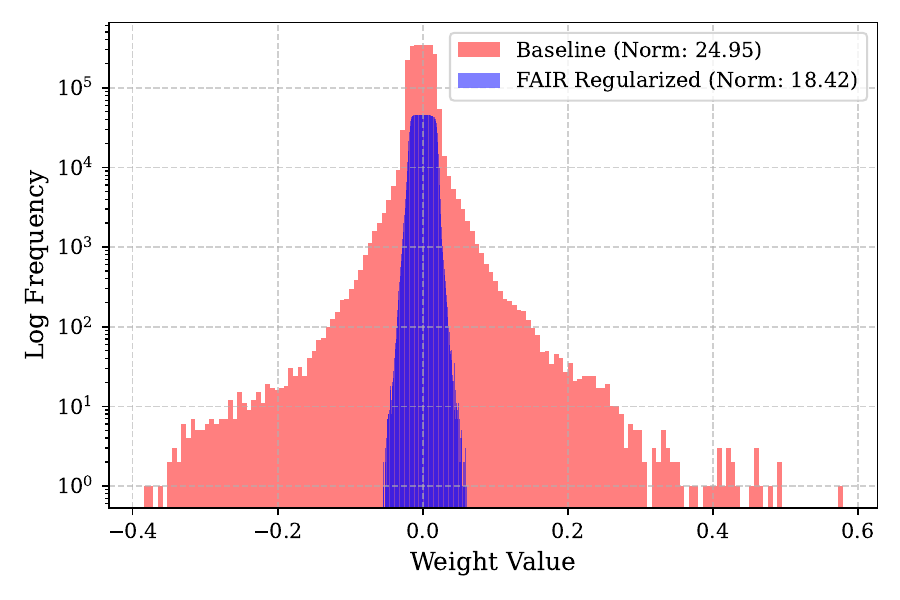}
        \caption{Weight distributions}
        \label{fig:weights_a}
    \end{subfigure}
    \begin{subfigure}{0.45\linewidth}
        \includegraphics[width=\linewidth]{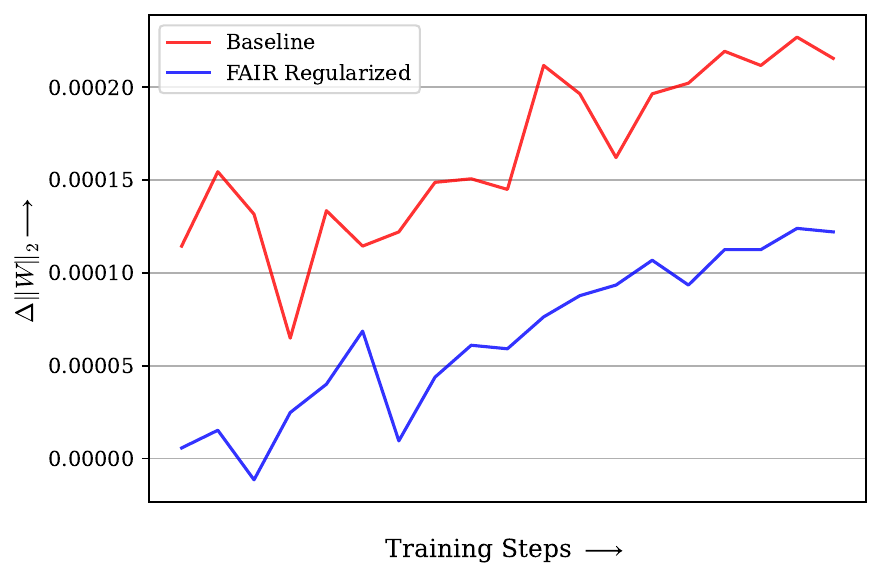}
        \caption{Weight norm update during training}
        \label{fig:weights_b}
    \end{subfigure}
    \caption{\textbf{Regularization Dynamics of FAIR.} (a) Weight distributions of the final classification layer demonstrate that FAIR compresses the hypothesis space, eliminating extreme outlier weights. (b) Tracking the change in weight magnitude ($\Delta \|W\|_2$) during training confirms that FAIR restrains parameter growth from the onset, indicating navigation of a smoother loss landscape.}
    \label{fig:weights}
\end{figure}

\subsubsection{Weight Distribution and Simplicity:} Effective regularization manifests as a constraint on weight magnitude. \cref{fig:weights_a} visualizes the weight distribution of the final classification layer for both the Base and FAIR-regularized models. The baseline model exhibits a heavy-tailed distribution with numerous extreme outlier weights, indicative of a highly non-linear, overfitted decision boundary. In contrast, FAIR compresses this distribution, substantially reducing the norm ($\|W\|_2$ drops from 24.95 to 18.42). This confirms FAIR successfully restricts the hypothesis space, enforcing a smoother, low-complexity decision boundary.

As shown in \cref{fig:weights_b}, the FAIR-regularized model exhibits a significantly lower and more stable increase in weight magnitude per step ($\Delta \|W\|_2$). Because weight updates are directly proportional to the loss gradient, this bounded trajectory confirms that the structural prior conditions the optimizer to navigate a fundamentally smoother, flatter loss landscape, preventing the aggressive parameter updates associated with memorizing high-frequency textural noise.

\begin{figure}[tb]
    \centering
    \includegraphics[width=0.5\linewidth]{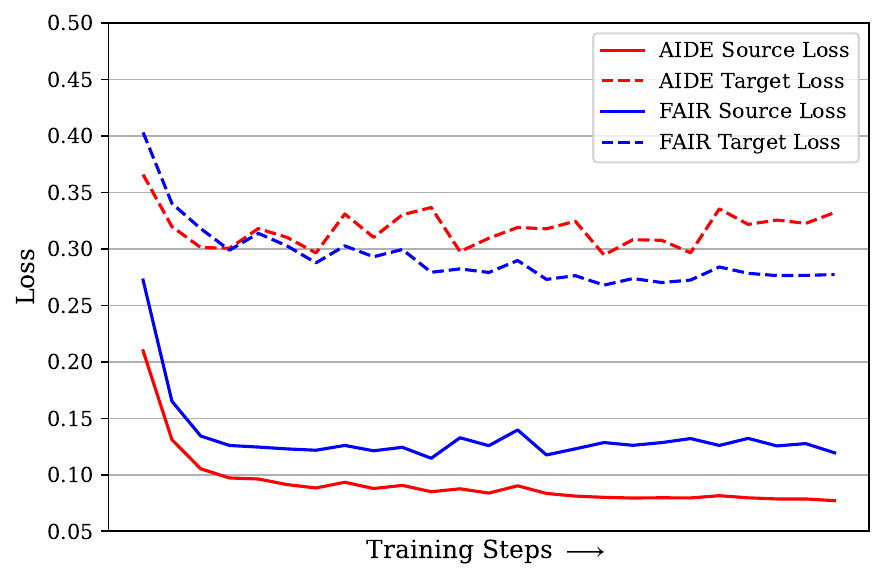}
    \caption{Source and target domain loss trajectories. FAIR imposes a slight source-domain penalty to stabilize and reduce the out-of-distribution target loss, exemplifying the bias-variance tradeoff. Note: Target loss curves were computed post-hoc by evaluating saved epoch checkpoints on a 5\% subset of the target domains. This target data was used strictly for visualization and remained completely unseen by the optimizer during training and hyperparameter selection.}
    \label{fig:overfitting_curves}
\end{figure}

\subsubsection{Loss Trajectory and Generalization Gap:} To explicitly visualize this reduction in source-domain overfitting, we conducted a post-hoc evaluation of the saved training checkpoints on a 5\% sample of the target domains to reconstruct the out-of-distribution loss trajectory (\cref{fig:overfitting_curves}). We emphasize that this evaluation is strictly analytical; target data was completely withheld during the actual optimization process.

The unregularized AIDE model exhibits a classic overfitting trajectory: it achieves an extremely low loss on the source domain while suffering from a high, volatile loss on the out-of-distribution target domains. Conversely, the FAIR training trajectory clearly demonstrates the anticipated bias-variance tradeoff. By constraining the optimization landscape, FAIR prevents the model from perfectly fitting the source domain (evidenced by a slightly higher source loss), but this penalty yields a significantly lower and more stable target loss. 

\begin{figure}[tb]
    \centering
    \includegraphics[width=0.5\linewidth]{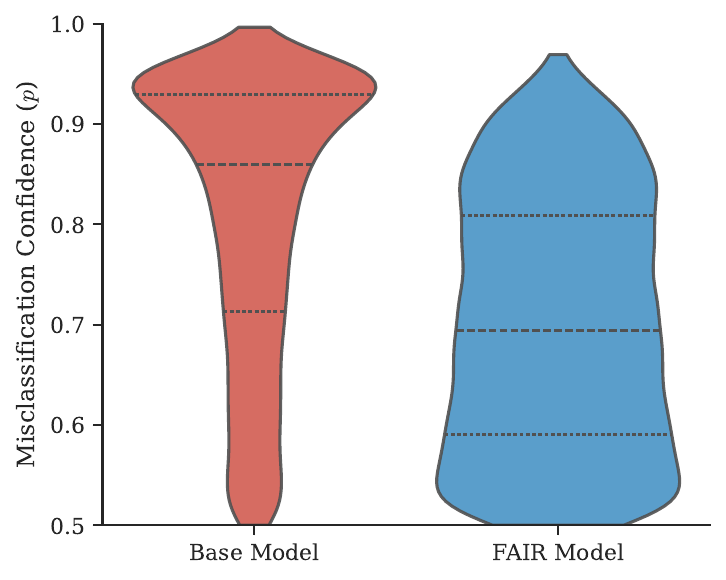}
    \caption{Global distribution of misclassification confidence: aggregated across GenImage, FAIR suppresses the base model's overconfident errors ($p > 0.9$), yielding a fundamentally softer and better-calibrated decision boundary.}
    \label{fig:margin_violin}
\end{figure}

\subsubsection{Misclassification Confidence and Calibration:} A hallmark of shortcut learning is that overfitted models become highly confident in their out-of-distrib\-ution errors. To evaluate the calibration of our smoothed decision boundary, we aggregated the prediction confidence (softmax probabilities) for every misclassified sample across the entire GenImage benchmark. As visualized in \cref{fig:margin_violin}, the baseline model exhibits a severely top-heavy error distribution, frequently committing to incorrect predictions with extreme, unjustified certainty ($p > 0.9$). Conversely, FAIR compresses this distribution, aggressively suppressing overconfident outliers and shifting the mass toward the uncertainty threshold ($p \approx 0.5$). This confirms that even when the smoothed structural boundary misclassifies a target sample, it exhibits a healthy, calibrated uncertainty, avoiding the overconfident failure modes characteristic of models that have memorized localized textural shortcuts.

\begin{table}[tb]
\centering
\caption{Centered Kernel Alignment (CKA) comparison on the SDv1.4 and ProGAN training sets. Standard base features exhibit high statistical dependence on the label ($Y$), whereas the isolated SCS features yield near-zero dependence.}
\label{tab:nhsic}
\resizebox{0.9\linewidth}{!}{
\setlength{\tabcolsep}{5pt}
\begin{tabular}{l | c c | c c}
\toprule
\textbf{CKA Score} & \multicolumn{2}{c|}{\textbf{SDv1.4 (Training Set)}} & \multicolumn{2}{c}{\textbf{ProGAN (Training Set)}} \\
\cmidrule(lr){2-3} \cmidrule(lr){4-5}
\textit{Lower is more invariant} & \textbf{Base (AIDE)} & \textbf{SCS (Ours)} & \textbf{Base (AIDE)} & \textbf{SCS (Ours)} \\
\midrule
Dependence on Label ($Y$) & 0.82 & \textbf{0.07} & 0.53 & \textbf{0.01} \\
\bottomrule
\end{tabular}
}
\end{table}

\subsubsection{Quantitative Independence of the Structural Prior:}
Finally, to mathematically verify that the structural prior ($X^*$) lacks the signal to act as a trivial discriminative shortcut ($I(X^*;Y) \approx 0$), we evaluated the Centered Kernel Alignment (CKA). Because CKA is a non-parametric measure bounded strictly to $[0, 1]$, it effectively calculates the strength of conditional dependence in high-dimensional spaces. As detailed in \cref{tab:nhsic}, the standard Base (AIDE) features exhibit high statistical dependence on the label. In contrast, our isolated SCS features yield near-zero dependence. This quantitatively confirms the structural prior contains virtually no predictive shortcut, forcing the network to anchor features to invariant geometry rather than bypassing the primary backbone.

\subsection{Robustness to Post-Processing: JPEG Compression}

In real-world deployment, AI-generated images are frequently subjected to transmission degradations, most notably JPEG compression. Because compression algorithms act as low-pass filters that actively destroy the high-frequency micro-textures, they provide a rigorous test for detectors that may be over-reliant on localized textural artifacts.

\begin{figure}[tb]
    \centering
    \includegraphics[width=0.55\linewidth]{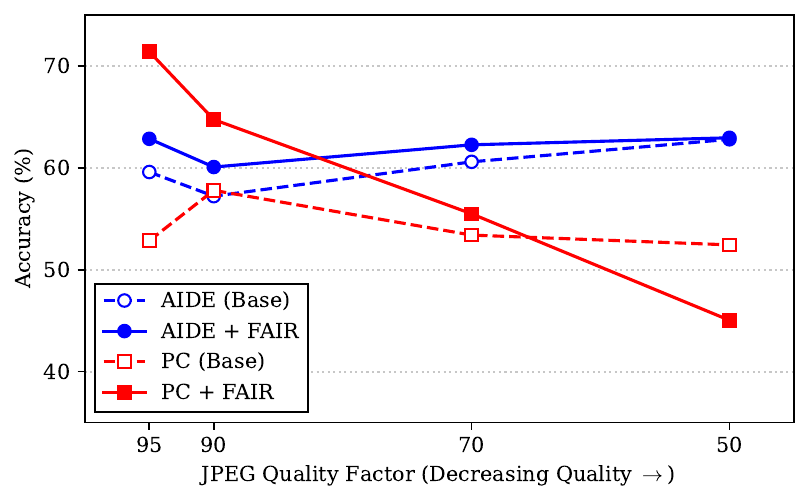}
    \caption{Robustness to JPEG Compression: accuracy decay curves under varying compression levels show FAIR significantly enhances robustness at mild-to-moderate compression qualities.}
    \label{fig:jpeg_robustness}
\end{figure}

To evaluate resilience, we subjected the GenImage test set to varying levels of JPEG compression. As visualized in \cref{fig:jpeg_robustness}, FAIR significantly improves resilience at mild to moderate compression levels (e.g., yielding an $+18.55\%$ gain for PatchCraft at $Q=95$). While AIDE+FAIR maintains superiority across all qualities, PatchCraft's performance drops below the baseline at extreme compression ($Q=50$). We attribute this to the specific nature of heavy JPEG degradation: at low qualities, compression introduces severe 8x8 blocking artifacts. Because FAIR explicitly regularizes the weights to anchor on structural geometry, the model becomes susceptible to these false artificial edges. Conversely, the unregularized baseline simply matches localized noise and is less directly deceived by global structural distortions. However, the overall trend confirms that anchoring decisions on macro-structural priors preserves robustness longer than relying strictly on unregularized local artifacts.

\begin{table}[htbp]
\centering
\caption{Comprehensive ablation of the AIDE baseline on GenImage. We compare our macro-structural prior (FAIR with SCS) against traditional parameter-level regularizers (Dropout, $L_1$, $L_2$) and alternative feature priors (Random, LPIPS, Sobel, Laplacian).}
\label{tab:comprehensive_ablation}
\resizebox{\linewidth}{!}{
\setlength{\tabcolsep}{2pt}
\begin{tabular}{l c |@{\hspace{6pt}} c c c c c c c @{\hspace{6pt}}|@{\hspace{6pt}} c c c c @{\hspace{6pt}}|@{\hspace{6pt}} c}
\toprule
\multirow{3}{*}{\textbf{Domains}} & \multirow{3}{*}{\textbf{\makecell{Base\\(AIDE)}}} & \multicolumn{7}{c@{\hspace{6pt}}|@{\hspace{6pt}}}{\textbf{Traditional Regularization}} & \multicolumn{4}{c@{\hspace{6pt}}|@{\hspace{6pt}}}{\textbf{Alternative Priors}} & \textbf{Proposed} \\
\cmidrule(lr){3-9} \cmidrule(lr){10-13} \cmidrule(lr){14-14}
& & \multirow{2}{*}{\textbf{Drop.}} & \multicolumn{3}{c}{\textbf{$L_1$ Sweep}} & \multicolumn{3}{c@{\hspace{6pt}}|@{\hspace{6pt}}}{\textbf{$L_2$ Sweep}} & \multirow{2}{*}{\textbf{Rand.}} & \multirow{2}{*}{\textbf{LPIPS}} & \multirow{2}{*}{\textbf{Sobel}} & \multirow{2}{*}{\textbf{Lapl.}} & \multirow{2}{*}{\textbf{SCS}} \\
\cmidrule(lr){4-6} \cmidrule(lr){7-9}
& & & \textbf{$10^{-4}$} & \textbf{$10^{-3}$} & \textbf{$10^{-2}$} & \textbf{$10^{-4}$} & \textbf{$10^{-3}$} & \textbf{$10^{-2}$} & & & & & \\
\midrule
Midjourney  & 79.4 & 76.5 & 71.6 & 72.3 & 50.0 & 77.0 & 74.9 & 71.7 & 77.0 & 71.6 & 73.2 & 75.1 & \textbf{83.6} \\
SDv1.4 (Src)& \textbf{99.7} & 99.7 & 99.1 & 98.8 & 50.0 & 99.7 & 99.6 & 99.1 & 99.7 & 99.3 & 99.5 & 99.6 & 99.6 \\
SDv1.5      & \textbf{99.8} & 99.6 & 99.1 & 98.9 & 50.0 & 99.7 & 99.6 & 99.1 & 99.7 & 99.1 & 99.4 & 99.5 & 99.6 \\
ADM         & 78.5 & 75.3 & 77.9 & 79.0 & 50.0 & 74.6 & 76.3 & 76.7 & 74.5 & 70.6 & 72.0 & 72.6 & \textbf{84.1} \\
GLIDE       & 91.8 & 90.7 & 86.6 & 92.0 & 50.0 & 92.0 & 89.7 & 87.2 & 90.8 & 85.9 & 85.8 & 88.4 & \textbf{96.7} \\
Wukong      & 98.7 & 98.4 & 97.0 & 96.6 & 50.0 & 98.6 & 98.1 & 96.9 & 98.6 & 96.3 & 97.1 & 97.5 & \textbf{99.1} \\
VQDM        & 80.3 & 81.7 & 82.0 & 81.7 & 50.0 & 81.6 & 82.3 & 80.6 & 81.9 & 77.4 & 79.0 & 79.1 & \textbf{87.2} \\
BigGAN      & 66.9 & 70.4 & 69.1 & 70.9 & 50.0 & 70.6 & 69.8 & 68.5 & 70.7 & 64.8 & 65.6 & 66.7 & \textbf{78.2} \\
\midrule
\textbf{Mean} & 86.9 & 86.5 & 85.3 & 86.3 & 50.0 & 86.7 & 86.3 & 85.0 & 86.6 & 83.1 & 83.9 & 84.8 & \textbf{91.0} \\
\bottomrule
\end{tabular}
}
\end{table}

\subsection{Ablation on Alternative Regularizations and Priors}

\subsubsection{FAIR vs. Traditional Regularization:} We compare FAIR against indiscriminate parameter-level regularizers (\cref{tab:comprehensive_ablation}). While the standard Dropout baseline (at $p=0.5$) offers a marginal improvement on radically different domains (e.g., BigGAN), it fails to improve the mean generalization. To verify if standard weight decay could solve this generalization gap, we applied full sweeps of $L_1$ and $L_2$ penalties directly to the loss. As the $L_2$ penalty scales from $10^{-4}$ to $10^{-2}$, mean cross-generator performance actively drops. Similarly, $L_1$ regularization forces harmful sparsity; at $10^{-2}$, it collapses the network entirely (50.00\% random chance). This confirms that indiscriminate regularizers only shrink weights or destroy features; they do not explicitly guide the network away from artifact shortcuts, a task uniquely solved by FAIR.

\subsubsection{SCS vs. Alternative Feature Priors:} The specific choice of the structural prior is also critical. In \cref{tab:comprehensive_ablation}, we substitute SCS with non-structural features (Random, LPIPS~\cite{zhang_unreasonable_2018}) and classical micro-structural priors (Sobel, Laplacian). Uniform Random features provide no meaningful semantic constraint, while LPIPS features are too low-level and texture-biased, causing the model to overfit and perform worse than the baseline. Furthermore, substituting SCS with pixel-level edge maps (Sobel, 1st-order gradient; Laplacian, 2nd-order derivative) causes noticeable performance drops. Because localized micro-structures are distorted differently by different generative engines, they serve as poor generalization anchors. SCS uniquely succeeds because it provides an orthogonal, macro-structural bottleneck that remains fundamentally consistent across diverse generators.

\begin{table}[htbp]
\centering
\caption{Ablation of the Base AIDE model demonstrating FAIR improves generalization even when primary feature streams (CLIP-based, $F_1$ or Patchwise, $F_2$) are omitted.}
\label{tab:ablation}
\resizebox{0.675\linewidth}{!}{
\begin{tabular}{l@{\hspace{10pt}}|@{\hspace{10pt}}cc@{\hspace{10pt}}|@{\hspace{10pt}}cc@{\hspace{10pt}}|@{\hspace{10pt}}cc}
\toprule
\multirow{2}{*}{Domains} & \multicolumn{2}{c@{\hspace{10pt}}|@{\hspace{10pt}}}{$F_1$ only} & \multicolumn{2}{c@{\hspace{10pt}}|@{\hspace{10pt}}}{$F_2$ only} & \multicolumn{2}{c}{Both} \\
& { Base } & { FAIR } & { Base } & { FAIR } & { Base } & { FAIR } \\
\midrule
Midjourney & 84.72 & \textbf{86.77} & 73.19 & \textbf{77.84} & 79.38 & \textbf{83.62} \\
SDv1.4 & \textbf{96.96} & 90.08 & 99.16 & \textbf{99.19} & \textbf{99.74} & 99.63 \\
SDv1.5 & \textbf{96.90} & 90.18 & 99.16 & 99.16 & \textbf{99.76} & 99.61 \\
ADM & 53.76 & \textbf{62.46} & 80.35 & \textbf{85.11} & 78.54 & \textbf{84.05} \\
Glide & 85.63 & \textbf{88.46} & 87.07 & \textbf{92.10} & 91.82 & \textbf{96.66} \\
Wukong & \textbf{92.92} & 89.82 & 97.37 & \textbf{97.93} & 98.65 & \textbf{99.11} \\
VQDM & 59.12 & \textbf{74.81} & 83.73 & \textbf{86.33} & 80.26 & \textbf{87.23} \\
BigGAN & 77.75 & \textbf{87.82} & 69.63 & \textbf{74.11} & 66.89 & \textbf{78.18} \\
\midrule
Mean & 80.97 & \textbf{83.80} & 86.20 & \textbf{88.97} & 86.88 & \textbf{91.01} \\
\bottomrule
\end{tabular}
}
\end{table}

\subsubsection{Feature Ablation and Decisiveness:} By constraining the weights, FAIR forces the network to utilize its available features more decisively. ~\cref{tab:ablation} shows an ablation study where AIDE's primary features, CLIP semantic ($F_1$) and Patchwise DCT ($F_2$), are independently excluded. In every scenario, applying FAIR improves the mean accuracy compared to the respective baseline. This indicates that FAIR successfully guides the model toward a generalized weight distribution, preventing the network from indiscriminately co-adapting to source-specific noise in either the semantic or spectral domains.

\section{Conclusion}
\label{sec:conclusion}
We introduced Feature-Augmented Implicit Regularization (FAIR), a novel strategy to combat source-domain overfitting and improve cross-generator generalization in AI-generated image detection. By introducing domain-invariant Scene Composition Structure (SCS) features as an orthogonal structural prior ($X^*$) strictly during training, FAIR geometrically constrains the optimizer to discover a smoother, lower-complexity decision boundary. Crucially, by intentionally discarding this prior at inference, FAIR embeds these structural constraints into the primary textural backbone with zero added architectural or computational overhead. Our method establishes new state-of-the-art robustness not only on standardized benchmarks like GenImage, but across massive, unseen in-the-wild diffusion manifolds, proving the critical advantage of informed geometric priors in representation learning.



%
%
\bibliographystyle{splncs04}
\bibliography{main}

@String(CVPR  = {IEEE Conf. Comput. Vis. Pattern Recog.})

@String(ICCV  = {Int. Conf. Comput. Vis.})

@String(ECCV  = {Eur. Conf. Comput. Vis.})

@String(ICIP  = {IEEE Int. Conf. Image Process.})

@String(ICASSP=	{ICASSP})

@String(CVPR  = {CVPR})

@String(ICCV  = {ICCV})

@String(ECCV  = {ECCV})

@String(ICIP  = {ICIP})

@inproceedings{rombach_high-resolution_2022,
    title = {High-{Resolution} {Image} {Synthesis} with {Latent} {Diffusion} {Models}},
    doi = {10.1109/CVPR52688.2022.01042},
    booktitle = {2022 {IEEE}/{CVF} {Conference} on {Computer} {Vision} and {Pattern} {Recognition} ({CVPR})},
    author = {Rombach, Robin and Blattmann, Andreas and Lorenz, Dominik and Esser, Patrick and Ommer, Björn},
    month = jun,
    year = {2022},
    pages = {10674--10685},
}

@inproceedings{yan_sanity_2025,
    title = {A {Sanity} {Check} for {AI}-generated {Image} {Detection}},
    author = {Yan, Shilin and Li, Ouxiang and Cai, Jiayin and Hao, Yanbin and Jiang, Xiaolong and Hu, Yao and Xie, Weidi},
    booktitle = {International Conference on Learning Representations},
    pages = {70702--70720},
    volume = {2025},
    year = {2025}
}

@article{zhong_patchcraft_2024,
    title = {{PatchCraft}: {Exploring} {Texture} {Patch} for {Efficient} {AI}-generated {Image} {Detection}},
    shorttitle = {{PatchCraft}},
    doi = {10.48550/arXiv.2311.12397},
    journal = {arXiv},
    author = {Zhong, Nan and Xu, Yiran and Li, Sheng and Qian, Zhenxing and Zhang, Xinpeng},
    month = mar,
    year = {2023},
}

@inproceedings{wang_dire_2023,
    address = {Paris, France},
    title = {{DIRE} for {Diffusion}-{Generated} {Image} {Detection}},
    copyright = {https://doi.org/10.15223/policy-029},
    isbn = {979-8-3503-0718-4},
    doi = {10.1109/ICCV51070.2023.02051},
    language = {en},
    booktitle = {2023 {IEEE}/{CVF} {International} {Conference} on {Computer} {Vision} ({ICCV})},
    publisher = {IEEE},
    author = {Wang, Zhendong and Bao, Jianmin and Zhou, Wengang and Wang, Weilun and Hu, Hezhen and Chen, Hong and Li, Houqiang},
    month = oct,
    year = {2023},
    pages = {22388--22398},
}

@inproceedings{wang_cnn-generated_2020,
    address = {Seattle, WA, USA},
    title = {{CNN}-{Generated} {Images} {Are} {Surprisingly} {Easy} to {Spot}… for {Now}},
    copyright = {https://ieeexplore.ieee.org/Xplorehelp/downloads/license-information/IEEE.html},
    isbn = {978-1-7281-7168-5},
    doi = {10.1109/CVPR42600.2020.00872},
    language = {en},
    booktitle = {2020 {IEEE}/{CVF} {Conference} on {Computer} {Vision} and {Pattern} {Recognition} ({CVPR})},
    publisher = {IEEE},
    author = {Wang, Sheng-Yu and Wang, Oliver and Zhang, Richard and Owens, Andrew and Efros, Alexei A.},
    month = jun,
    year = {2020},
    pages = {8692--8701},
}

@inproceedings{frank_leveraging_2020,
    title = {Leveraging {Frequency} {Analysis} for {Deep} {Fake} {Image} {Recognition}},
    language = {en},
    booktitle = {Proceedings of the 37th {International} {Conference} on {Machine} {Learning}},
    publisher = {PMLR},
    author = {Frank, Joel and Eisenhofer, Thorsten and Schönherr, Lea and Fischer, Asja and Kolossa, Dorothea and Holz, Thorsten},
    month = nov,
    year = {2020},
    pages = {3247--3258},
}

@inproceedings{ju_fusing_2022,
    title = {Fusing {Global} and {Local} {Features} for {Generalized} {AI}-{Synthesized} {Image} {Detection}},
    doi = {10.1109/ICIP46576.2022.9897820},
    booktitle = {2022 {IEEE} {International} {Conference} on {Image} {Processing} ({ICIP})},
    author = {Ju, Yan and Jia, Shan and Ke, Lipeng and Xue, Hongfei and Nagano, Koki and Lyu, Siwei},
    month = oct,
    year = {2022},
    pages = {3465--3469},
}

@inproceedings{liu_detecting_2022,
    address = {Berlin, Heidelberg},
    title = {Detecting {Generated} {Images} by {Real} {Images}},
    isbn = {978-3-031-19780-2},
    doi = {10.1007/978-3-031-19781-9_6},
    booktitle = {Computer {Vision} – {ECCV} 2022: 17th {European} {Conference}, {Tel} {Aviv}, {Israel}, {October} 23–27, 2022, {Proceedings}, {Part} {XIV}},
    publisher = {Springer-Verlag},
    author = {Liu, Bo and Yang, Fan and Bi, Xiuli and Xiao, Bin and Li, Weisheng and Gao, Xinbo},
    month = oct,
    year = {2022},
    pages = {95--110},
}

@inproceedings{ojha_towards_2023,
    address = {Vancouver, BC, Canada},
    title = {Towards {Universal} {Fake} {Image} {Detectors} that {Generalize} {Across} {Generative} {Models}},
    copyright = {https://doi.org/10.15223/policy-029},
    isbn = {979-8-3503-0129-8},
    doi = {10.1109/CVPR52729.2023.02345},
    language = {en},
    booktitle = {2023 {IEEE}/{CVF} {Conference} on {Computer} {Vision} and {Pattern} {Recognition} ({CVPR})},
    publisher = {IEEE},
    author = {Ojha, Utkarsh and Li, Yuheng and Lee, Yong Jae},
    month = jun,
    year = {2023},
    pages = {24480--24489},
}

@INPROCEEDINGS{haque_novel_2025,
  author={Haque, Md Redwanul and Murshed, Manzur and Paul, Manoranjan and Lee, Tsz-Kwan},
  booktitle={2025 IEEE International Conference on Image Processing Workshops (ICIPW)}, 
  title={A Novel Image Similarity Metric For Scene Composition Structure}, 
  doi = {10.1109/ICIPW68931.2025.11386242},
  year={2025},
  volume={},
  number={},
  pages={446-451},
}

@article{zhu_gendet_2023,
    title = {{GenDet}: {Towards} {Good} {Generalizations} for {AI}-{Generated} {Image} {Detection}},
    shorttitle = {{GenDet}},
    doi = {10.48550/arXiv.2312.08880},
    journal = {arXiv},
    author = {Zhu, Mingjian and Chen, Hanting and Huang, Mouxiao and Li, Wei and Hu, Hailin and Hu, Jie and Wang, Yunhe},
    month = dec,
    year = {2023},
}

@inproceedings{zhu_genimage_2023,
    title = {{GenImage}: {A} {Million}-{Scale} {Benchmark} for {Detecting} {AI}-{Generated} {Image}},
    volume = {36},
    shorttitle = {{GenImage}},
    language = {en},
    booktitle = {Advances in Neural Information Processing Systems},
    author = {Zhu, Mingjian and Chen, Hanting and Yan, Qiangyu and Huang, Xudong and Lin, Guanyu and Li, Wei and Tu, Zhijun and Hu, Hailin and Hu, Jie and Wang, Yunhe},
    month = dec,
    year = {2023},
    pages = {77771--77782},
}

@inproceedings{ho_denoising_2020,
    title = {Denoising {Diffusion} {Probabilistic} {Models}},
    volume = {33},
    booktitle = {Advances in {Neural} {Information} {Processing} {Systems}},
    publisher = {Curran Associates, Inc.},
    author = {Ho, Jonathan and Jain, Ajay and Abbeel, Pieter},
    year = {2020},
    pages = {6840--6851},
}

@article{ahmed_discrete_1974,
    title = {Discrete {Cosine} {Transform}},
    volume = {C-23},
    issn = {1557-9956},
    doi = {10.1109/T-C.1974.223784},
    number = {1},
    journal = {IEEE Transactions on Computers},
    author = {Ahmed, Nasir and Natarajan, T. and Rao, Kamisetty R.},
    month = jan,
    year = {1974},
    pages = {90--93},
}

@inproceedings{he_deep_2016,
    title = {Deep {Residual} {Learning} for {Image} {Recognition}},
    doi = {10.1109/CVPR.2016.90},
    booktitle = {2016 {IEEE} {Conference} on {Computer} {Vision} and {Pattern} {Recognition} ({CVPR})},
    author = {He, Kaiming and Zhang, Xiangyu and Ren, Shaoqing and Sun, Jian},
    month = jun,
    year = {2016},
    pages = {770--778},
}

@inproceedings{touvron_training_2021,
    title = {Training data-efficient image transformers \& distillation through attention},
    language = {en},
    booktitle = {Proceedings of the 38th {International} {Conference} on {Machine} {Learning}},
    publisher = {PMLR},
    author = {Touvron, Hugo and Cord, Matthieu and Douze, Matthijs and Massa, Francisco and Sablayrolles, Alexandre and Jegou, Herve},
    month = jul,
    year = {2021},
    pages = {10347--10357},
}

@inproceedings{liu_swin_2021,
    address = {Montreal, QC, Canada},
    title = {Swin {Transformer}: {Hierarchical} {Vision} {Transformer} using {Shifted} {Windows}},
    copyright = {https://doi.org/10.15223/policy-029},
    isbn = {978-1-6654-2812-5},
    shorttitle = {Swin {Transformer}},
    doi = {10.1109/ICCV48922.2021.00986},
    language = {en},
    booktitle = {2021 {IEEE}/{CVF} {International} {Conference} on {Computer} {Vision} ({ICCV})},
    publisher = {IEEE},
    author = {Liu, Ze and Lin, Yutong and Cao, Yue and Hu, Han and Wei, Yixuan and Zhang, Zheng and Lin, Stephen and Guo, Baining},
    month = oct,
    year = {2021},
    pages = {9992--10002},
}

@inproceedings{tan_learning_2023,
    address = {Vancouver, BC, Canada},
    title = {Learning on {Gradients}: {Generalized} {Artifacts} {Representation} for {GAN}-{Generated} {Images} {Detection}},
    copyright = {https://doi.org/10.15223/policy-029},
    isbn = {979-8-3503-0129-8},
    shorttitle = {Learning on {Gradients}},
    doi = {10.1109/CVPR52729.2023.01165},
    language = {en},
    booktitle = {2023 {IEEE}/{CVF} {Conference} on {Computer} {Vision} and {Pattern} {Recognition} ({CVPR})},
    publisher = {IEEE},
    author = {Tan, Chuangchuang and Zhao, Yao and Wei, Shikui and Gu, Guanghua and Wei, Yunchao},
    month = jun,
    year = {2023},
    pages = {12105--12114},
}

@inproceedings{tan_rethinking_2024,
    address = {Seattle, WA, USA},
    title = {Rethinking the {Up}-{Sampling} {Operations} in {CNN}-{Based} {Generative} {Network} for {Generalizable} {Deepfake} {Detection}},
    copyright = {https://doi.org/10.15223/policy-029},
    isbn = {979-8-3503-5300-6},
    doi = {10.1109/CVPR52733.2024.02657},
    language = {en},
    booktitle = {2024 {IEEE}/{CVF} {Conference} on {Computer} {Vision} and {Pattern} {Recognition} ({CVPR})},
    publisher = {IEEE},
    author = {Tan, Chuangchuang and Liu, Huan and Zhao, Yao and Wei, Shikui and Gu, Guanghua and Liu, Ping and Wei, Yunchao},
    month = jun,
    year = {2024},
    pages = {28130--28139},
}

@inproceedings{radford_learning_2021,
    title = {Learning {Transferable} {Visual} {Models} {From} {Natural} {Language} {Supervision}},
    language = {en},
    booktitle = {Proceedings of the 38th {International} {Conference} on {Machine} {Learning}},
    publisher = {PMLR},
    author = {Radford, Alec and Kim, Jong Wook and Hallacy, Chris and Ramesh, Aditya and Goh, Gabriel and Agarwal, Sandhini and Sastry, Girish and Askell, Amanda and Mishkin, Pamela and Clark, Jack and Krueger, Gretchen and Sutskever, Ilya},
    month = jul,
    year = {2021},
    pages = {8748--8763},
}

@incollection{qian_thinking_2020,
    address = {Cham},
    title = {Thinking in {Frequency}: {Face} {Forgery} {Detection} by {Mining} {Frequency}-{Aware} {Clues}},
    volume = {12357},
    isbn = {978-3-030-58609-6 978-3-030-58610-2},
    shorttitle = {Thinking in {Frequency}},
    language = {en},
    booktitle = {Computer {Vision} – {ECCV} 2020},
    publisher = {Springer International Publishing},
    author = {Qian, Yuyang and Yin, Guojun and Sheng, Lu and Chen, Zixuan and Shao, Jing},
    editor = {Vedaldi, Andrea and Bischof, Horst and Brox, Thomas and Frahm, Jan-Michael},
    year = {2020},
    doi = {10.1007/978-3-030-58610-2_6},
    pages = {86--103},
}

@inproceedings{liu_global_2020,
    address = {Seattle, WA, USA},
    title = {Global {Texture} {Enhancement} for {Fake} {Face} {Detection} in the {Wild}},
    copyright = {https://ieeexplore.ieee.org/Xplorehelp/downloads/license-information/IEEE.html},
    isbn = {978-1-7281-7168-5},
    doi = {10.1109/CVPR42600.2020.00808},
    language = {en},
    booktitle = {2020 {IEEE}/{CVF} {Conference} on {Computer} {Vision} and {Pattern} {Recognition} ({CVPR})},
    publisher = {IEEE},
    author = {Liu, Zhengzhe and Qi, Xiaojuan and Torr, Philip H.S.},
    month = jun,
    year = {2020},
    pages = {8057--8066},
}

@article{zhang_spec_2019,
  title={Detecting and Simulating Artifacts in {GAN} Fake Images},
  author={Xu Zhang and Svebor Karaman and Shih-Fu Chang},
  journal={2019 IEEE International Workshop on Information Forensics and Security (WIFS)},
  year={2019},
  pages={1-6},
}

@misc{hendrycks2023gaussianerrorlinearunits,
      title={Gaussian Error Linear Units ({GELUs})}, 
      author={Dan Hendrycks and Kevin Gimpel},
      year={2016},
      eprint={1606.08415},
      archivePrefix={arXiv},
      primaryClass={cs.LG},
      url={https://arxiv.org/abs/1606.08415}, 
}

@misc{midjourney_2022,
    author = {{Midjourney, Inc.}},
    title = {Midjourney},
    howpublished = {Online},
    year = {2022},
    url = {https://www.midjourney.com},
    urldate = {2025-09-24},
    note = {{Accessed}: 2025-09-24}
}

@ARTICLE{fridrich_srm_2012,
    author={Fridrich, Jessica and Kodovsky, Jan},
    journal={IEEE Transactions on Information Forensics and Security}, 
    title={Rich Models for Steganalysis of Digital Images}, 
    year={2012},
    volume={7},
    number={3},
    pages={868-882},
    doi={10.1109/TIFS.2012.2190402}
}

@article{dropout_srivastava,
  author  = {Nitish Srivastava and Geoffrey Hinton and Alex Krizhevsky and Ilya Sutskever and Ruslan Salakhutdinov},
  title   = {Dropout: A Simple Way to Prevent Neural Networks from Overfitting},
  journal = {Journal of Machine Learning Research},
  year    = {2014},
  volume  = {15},
  number  = {56},
  pages   = {1929--1958},
  url     = {http://jmlr.org/papers/v15/srivastava14a.html}
}

@inproceedings{cortes2012l2,
author = {Cortes, Corinna and Mohri, Mehryar and Rostamizadeh, Afshin},
title = {L2 regularization for learning kernels},
year = {2009},
isbn = {9780974903958},
publisher = {AUAI Press},
address = {Arlington, Virginia, USA},
booktitle = {Proceedings of the Twenty-Fifth Conference on Uncertainty in Artificial Intelligence},
pages = {109–116},
numpages = {8},
location = {Montreal, Quebec, Canada},
series = {UAI '09}
}

@inproceedings{gan_goodfellow_nips,
 author = {Goodfellow, Ian J. and Pouget-Abadie, Jean and Mirza, Mehdi and Xu, Bing and Warde-Farley, David and Ozair, Sherjil and Courville, Aaron and Bengio, Yoshua},
 booktitle = {Advances in Neural Information Processing Systems},
 editor = {Z. Ghahramani and M. Welling and C. Cortes and N. Lawrence and K.Q. Weinberger},
 pages = {},
 publisher = {Curran Associates, Inc.},
 title = {Generative Adversarial Nets},
 volume = {27},
 year = {2014}
}

@article{healy_uniform_2024,
	title = {Uniform manifold approximation and projection},
	volume = {4},
	issn = {2662-8449},
	doi = {10.1038/s43586-024-00363-x},
	number = {1},
	journal = {Nature Reviews Methods Primers},
	author = {Healy, John and McInnes, Leland},
	month = nov,
	year = {2024},
	pages = {82},
}

@ARTICLE{svm1998,
  author={Hearst, M.A. and Dumais, S.T. and Osuna, E. and Platt, J. and Scholkopf, B.},
  journal={IEEE Intelligent Systems and their Applications}, 
  title={Support vector machines}, 
  year={1998},
  volume={13},
  number={4},
  pages={18-28},
  doi={10.1109/5254.708428}
}

@inproceedings{zhang_unreasonable_2018,
    title = {The {Unreasonable} {Effectiveness} of {Deep} {Features} as a {Perceptual} {Metric}},
    booktitle = {2018 {IEEE}/{CVF} {Conference} on {Computer} {Vision} and {Pattern} {Recognition}},
    author = {Zhang, Richard and Isola, Phillip and Efros, Alexei A. and Shechtman, Eli and Wang, Oliver},
    month = jun,
    year = {2018},
    pages = {586--595},
}

@article{vapnik2009new,
  title={A new learning paradigm: Learning using privileged information},
  author={Vapnik, Vladimir and Vashist, Akshay},
  journal={Neural networks},
  volume={22},
  number={5-6},
  pages={544--557},
  year={2009},
  publisher={Elsevier}
}

@misc{arjovsky2020invariantriskminimization,
      title={Invariant Risk Minimization}, 
      author={Martin Arjovsky and Léon Bottou and Ishaan Gulrajani and David Lopez-Paz},
      year={2019},
      eprint={1907.02893},
      archivePrefix={arXiv},
      primaryClass={stat.ML},
      url={https://arxiv.org/abs/1907.02893}, 
}

@article{geirhos_shortcut_2020,
	title = {Shortcut learning in deep neural networks},
	volume = {2},
	issn = {2522-5839},
	doi = {10.1038/s42256-020-00257-z},
	number = {11},
	journal = {Nature Machine Intelligence},
	author = {Geirhos, Robert and Jacobsen, Jörn-Henrik and Michaelis, Claudio and Zemel, Richard and Brendel, Wieland and Bethge, Matthias and Wichmann, Felix A.},
	month = nov,
	year = {2020},
	pages = {665--673},
}

@article{ganin_domain-adversarial_2016,
	title = {Domain-{Adversarial} {Training} of {Neural} {Networks}},
	volume = {17},
	url = {http://jmlr.org/papers/v17/15-239.html},
	number = {59},
	journal = {Journal of Machine Learning Research},
	author = {Ganin, Yaroslav and Ustinova, Evgeniya and Ajakan, Hana and Germain, Pascal and Larochelle, Hugo and Laviolette, François and March, Mario and Lempitsky, Victor},
	year = {2016},
	pages = {1--35},
}

@article{ben-david_theory_2010,
	title = {A theory of learning from different domains},
	volume = {79},
	issn = {1573-0565},
	doi = {10.1007/s10994-009-5152-4},
	number = {1},
	journal = {Machine Learning},
	author = {Ben-David, Shai and Blitzer, John and Crammer, Koby and Kulesza, Alex and Pereira, Fernando and Vaughan, Jennifer Wortman},
	month = may,
	year = {2010},
	pages = {151--175},
}

@InProceedings{pmlr-v235-chen24ay,
  title = 	 {{DRCT}: Diffusion Reconstruction Contrastive Training towards Universal Detection of Diffusion Generated Images},
  author =       {Chen, Baoying and Zeng, Jishen and Yang, Jianquan and Yang, Rui},
  booktitle = 	 {Proceedings of the 41st International Conference on Machine Learning},
  pages = 	 {7621--7639},
  year = 	 {2024},
  editor = 	 {Salakhutdinov, Ruslan and Kolter, Zico and Heller, Katherine and Weller, Adrian and Oliver, Nuria and Scarlett, Jonathan and Berkenkamp, Felix},
  volume = 	 {235},
  series = 	 {Proceedings of Machine Learning Research},
  month = 	 {21--27 Jul},
  publisher =    {PMLR},
  url = 	 {https://proceedings.mlr.press/v235/chen24ay.html},
}

@inproceedings{lu2023seeing_fake2m,
  title={Seeing is not always believing: Benchmarking human and model perception of {AI}-generated images},
  author={Lu, Zeyu and Huang, Di and Bai, Lei and Qu, Jingjing and Wu, Chengyue and Liu, Xihui and Ouyang, Wanli},
  booktitle={Advances in Neural Information Processing Systems},
  volume={36},
  pages={25435--25447},
  year={2023}
}

@inproceedings{corvi2023detection,
  title={On the detection of synthetic images generated by diffusion models},
  author={Corvi, Riccardo and Cozzolino, Davide and Zingarini, Giada and Siracusa, Giovanni and Verdoliva, Luisa},
  booktitle={2023 IEEE International Conference on Acoustics, Speech and Signal Processing (ICASSP)},
  pages={1--5},
  year={2023},
  organization={IEEE}
}

\clearpage
\appendix
\renewcommand{\thetable}{S\arabic{table}}
\renewcommand{\thefigure}{S\arabic{figure}}
\renewcommand{\thelstlisting}{S\arabic{lstlisting}}

\section{Implementation Details}

\subsection{PyTorch Code}

To demonstrate the straightforward, plug-and-play nature of FAIR, we provide the core PyTorch implementation in \cref{lst:fair_code}. FAIR acts as a lightweight training wrapper around any base deepfake detector, augmenting only the first classifier layer to accept the projected structural prior. Crucially, the \texttt{save\_model} method illustrates how the augmented weights are sliced prior to export / checkpoint saving. This produces a standard checkpoint that perfectly matches the architecture of the unmodified base detector, ensuring zero additional computational overhead during deployment.

\begin{lstlisting}[language=Python, caption={PyTorch implementation of the FAIR training wrapper and zero-overhead export.}, label={lst:fair_code}]
import torch
import torch.nn as nn

from base_detector import BaseDetector

class FAIRWrapper(nn.Module):
    def __init__(self, N_prior, M_prior):
        super().__init__()
        self.base = BaseDetector()
        # has a feature_extractor and a classifier
        
        self.prior_projector = nn.Sequential(
            nn.Linear(N_prior, M_prior),
            nn.GELU()
        )
        
        old_layer1 = self.base.classifier.layer1
        self.D = old_layer1.in_features
        
        self.base.classifier.layer1 = nn.Linear(
            self.D + M_prior, old_layer1.out_features
        )

    def forward_train(self, image, scs):
        X = self.base.feature_extractor(image)
        
        X_star = self.prior_projector(scs)
        X_aug = torch.cat([X, X_star], dim=1)
        
        return self.base.classifier(X_aug)

    def save_model(self, save_path):
        state_dict = self.base.state_dict()

        weight_key = 'classifier.layer1.weight'
        w = state_dict[weight_key]
        state_dict[weight_key] = w[:, :self.D]

        torch.save(state_dict, save_path)
\end{lstlisting}

\subsection{FAIR Hyperparameters}

As detailed in the main text, the core training configurations (e.g., batch size, learning rate) follow the default settings of the respective base models, Patch\-Craft~\cite{zhong_patchcraft_2024} and AIDE~\cite{yan_sanity_2025}. This ensures a rigorous and fair baseline comparison. ~\cref{tab:fair_hyperparams} outlines the hyperparameters exclusively introduced by our regularization module: the raw Scene Composition Structure (SCS) feature dimension ($N$) and the compact projected prior dimension ($M$).

\begin{table}[htbp]
    \centering
    \caption{FAIR-specific architectural hyperparameters across base methods and training benchmarks.}
    \label{tab:fair_hyperparams}
    \begin{tabular}{l@{\hspace{10pt}}l@{\hspace{10pt}}c@{\hspace{10pt}}c}
        \toprule
        \textbf{\makecell[l]{Base\\Method}} & \textbf{\makecell[l]{Training\\Benchmark}} & \textbf{\makecell{Raw\\Prior \boldmath$N$}} & \textbf{\makecell{Projected\\Prior \boldmath$M$}} \\
        \midrule
        \multirow{2}{*}{AIDE} & GenImage & 1024 & 256 \\
        & AIGCDetect & 256 & 64 \\
        \midrule
        \multirow{2}{*}{PatchCraft} & GenImage & 1024 & 32 \\
        & AIGCDetect & 1024 & 32 \\
        \bottomrule
    \end{tabular}
\end{table}

\clearpage

\section{Detailed Benchmark Results}

\subsection{UnivFD Benchmark}

\cref{tab:univfd_detailed} provides the comprehensive domain-by-domain performance breakdown on the UnivFD~\cite{ojha_towards_2023} benchmark. Consistent with the aggregated results presented in the main text, applying the FAIR regularization head yields targeted improvements across a diverse range of unseen GAN and diffusion architectures.

\begin{table}[htbp]
\centering
\caption{Detailed In-the-Wild Cross-Benchmark Transfer on the UnivFD~\cite{ojha_towards_2023} Benchmark. Models are trained on GenImage~\cite{zhu_genimage_2023} SDv1.4 and tested zero-shot across 21 diverse GAN and Diffusion domains. Bold values indicate an improvement by FAIR over its respective base architecture.}
\label{tab:univfd_detailed}
\resizebox{0.8\textwidth}{!}{
\begin{tabular}{l|@{\hspace{5pt}}cc|@{\hspace{5pt}}c@{\hspace{18pt}}c}
\toprule
\multirow{2}{*}{\textbf{Domain}} & \multicolumn{2}{c@{\hspace{5pt}}|@{\hspace{5pt}}}{\textbf{PatchCraft (PC)}} & \multicolumn{2}{c}{\textbf{AIDE}} \\
& \textbf{Base} & \textbf{FAIR} & \textbf{Base} & \textbf{FAIR} \\
\midrule
BigGAN & 95.90 & 93.00 & 77.20 & 73.38 \\
CRN & 50.02 & 50.00 & 51.82 & 44.42 \\
DeepFake & 57.95 & 56.97 & 54.47 & 52.45 \\
GauGAN & 72.60 & \textbf{84.14} & 64.36 & \textbf{65.44} \\
IMLE & 50.02 & 50.00 & 68.05 & \textbf{73.61} \\
SAN & 87.44 & \textbf{87.67} & 58.45 & \textbf{65.07} \\
StarGAN & 88.33 & 79.17 & 80.27 & \textbf{86.87} \\
WhichFaceIsReal & 99.95 & 99.47 & 74.55 & \textbf{79.05} \\
CycleGAN & 88.05 & 86.05 & 74.39 & \textbf{81.93} \\
ProGAN & 69.02 & \textbf{79.09} & 72.75 & 71.50 \\
StyleGAN & 100.0 & 100.0 & 70.05 & \textbf{75.35} \\
StyleGAN2 & 92.28 & 90.72 & 80.33 & \textbf{82.12} \\
SeeingDark & 99.98 & 99.43 & 64.72 & 63.33 \\
DALL-E & 88.95 & \textbf{90.35} & 87.65 & \textbf{93.10} \\
GLIDE (100-10) & 89.35 & \textbf{92.30} & 93.00 & \textbf{97.15} \\
GLIDE (100-27) & 82.35 & \textbf{88.80} & 92.95 & \textbf{96.70} \\
GLIDE (50-27) & 86.25 & \textbf{90.35} & 93.70 & \textbf{96.40} \\
Guided & 80.95 & \textbf{81.10} & 70.95 & \textbf{80.35} \\
LDM (100) & 94.70 & 93.65 & 97.30 & \textbf{98.65} \\
LDM (200) & 94.80 & 93.60 & 97.20 & \textbf{98.90} \\
LDM (200-cfg) & 93.40 & 92.60 & 97.80 & \textbf{98.25} \\
\midrule
\textbf{Mean} & 83.92 & \textbf{84.69} & 77.24 & \textbf{79.72} \\
\bottomrule
\end{tabular}
}
\end{table}

\clearpage

\subsection{Fake2M Benchmark}

\cref{tab:fake2m_detailed} details the zero-shot transfer performance across all domains within the Fake2M dataset. The results further validate that anchoring weights on macro-structural priors via FAIR prevents severe degradation on complex, unseen generator distributions.

\begin{table}[htbp]
\centering
\caption{Detailed In-the-Wild Cross-Benchmark Transfer on the Fake2M~\cite{lu2023seeing_fake2m} Benchmark. Models are trained on GenImage SDv1.4 and tested zero-shot. Bold values indicate an improvement by FAIR over its respective base architecture.}
\label{tab:fake2m_detailed}
\resizebox{0.9\textwidth}{!}{
\begin{tabular}{l|@{\hspace{5pt}}cc|@{\hspace{5pt}}c@{\hspace{20pt}}c}
\toprule
\multirow{2}{*}{\textbf{Domain}} & \multicolumn{2}{c@{\hspace{5pt}}|@{\hspace{5pt}}}{\textbf{PatchCraft (PC)}} & \multicolumn{2}{c}{\textbf{AIDE}} \\
& \textbf{Base} & \textbf{FAIR} & \textbf{Base} & \textbf{FAIR} \\
\midrule
Midjourney v5 & 23.66 & 21.99 & 15.16 & 11.29 \\
SDv1.5 (DPMSolver-25) & 96.37 & \textbf{97.88} & 99.67 & \textbf{99.85} \\
SDv1.5R (DPMSolver-25) & 98.52 & \textbf{99.51} & 98.29 & \textbf{98.69} \\
CogView2 & 96.84 & 76.97 & 26.05 & \textbf{51.04} \\
SDv2.1 (DPMSolver-25) & 85.82 & \textbf{97.59} & 88.95 & \textbf{91.88} \\
IF (DPMSolver++-25) & 72.12 & 56.01 & 60.12 & \textbf{85.76} \\
IF (DPMSolver++-50) & 71.08 & 54.92 & 59.72 & \textbf{86.81} \\
IF (DDIM-25) & 53.85 & \textbf{62.07} & 67.76 & \textbf{68.78} \\
IF (DDPM-50) & 55.28 & \textbf{58.06} & 78.18 & \textbf{80.11} \\
IF (DPMSolver++-10) & 78.64 & 70.83 & 72.72 & \textbf{90.63} \\
IF (DDIM-50) & 40.36 & \textbf{49.19} & 60.69 & \textbf{61.90} \\
StyleGAN3 (t-AFHQ) & 99.76 & 98.49 & 61.65 & \textbf{67.52} \\
StyleGAN3 (r-FFHQ) & 100.0 & 100.0 & 82.32 & 80.00 \\
StyleGAN3 (t-MetFaces) & 98.07 & \textbf{99.33} & 75.11 & \textbf{82.37} \\
StyleGAN3 (t-FFHQ) & 100.0 & 100.0 & 84.40 & 79.73 \\
StyleGAN3 (r-AFHQ) & 99.17 & 97.60 & 64.61 & \textbf{71.09} \\
StyleGAN3 (r-MetFaces) & 98.34 & \textbf{99.46} & 79.72 & \textbf{87.68} \\
\midrule
\textbf{Mean} & 80.46 & 78.82 & 69.12 & \textbf{76.18} \\
\bottomrule
\end{tabular}
}
\end{table}

\clearpage

\subsection{AIGCDetect Benchmark}

\cref{tab:aigcdetect_results} presents the detailed performance evaluation on the standardized AIGC\-Detect~\cite{zhong_patchcraft_2024} benchmark. In this setting, models are trained exclusively on ProGAN and evaluated zero-shot across a diverse suite of 16 generative models, encompassing both legacy GANs and modern diffusion architectures. As is characteristic of any regularization method, FAIR introduces a specific inductive bias towards macro-structural features; consequently, while it significantly elevates aggregate generalizability, it may result in minor performance trade-offs on certain isolated domains rather than uniform gains across every individual distribution.

\begin{table}[htbp]
\caption{Performance comparison of the proposed FAIR method against the base detector and existing detectors on the AIGCDetect~\cite{zhong_patchcraft_2024} benchmark datasets, where the best and second-best results in each column are marked in \textbf{bold} and \underline{underline}, respectively.}
\label{tab:aigcdetect_results}
\centering
\resizebox{\textwidth}{!}{
\begin{tabular}{lccccccccccccccccc|c}
\toprule
\textbf{Method} & \rotatebox{80}{\textbf{ProGAN}} & \rotatebox{80}{\textbf{StyleGAN}} & \rotatebox{80}{\textbf{BigGAN}} & \rotatebox{80}{\textbf{CycleGAN}} & \rotatebox{80}{\textbf{StarGAN}} & \rotatebox{80}{\textbf{GauGAN}} & \rotatebox{80}{\textbf{StyleGAN2}} & \rotatebox{80}{\textbf{WFIR}} & \rotatebox{80}{\textbf{ADM}} & \rotatebox{80}{\textbf{Glide}} & \rotatebox{80}{\textbf{Midjourney}} & \rotatebox{80}{\textbf{SD v1.4}} & \rotatebox{80}{\textbf{SD v1.5}} & \rotatebox{80}{\textbf{VQDM}} & \rotatebox{80}{\textbf{Wukong}} & \rotatebox{80}{\textbf{DALLE2}} & \rotatebox{80}{\textbf{SDXL}} & \rotatebox{80}{\textbf{Mean}} \\
\midrule
CNNSpot~\cite{wang_cnn-generated_2020} & \textbf{ 100.0 } & { 90.17 } & { 71.17 } & { 87.62 } & { 94.60 } & { 81.42 } & { 86.91 } & { 91.65 } & { 60.39 } & { 58.07 } & { 51.39 } & { 50.57 } & { 50.53 } & { 56.46 } & { 51.03 } & { 50.45 } & { 53.03 } & { 69.73 } \\
FreDect~\cite{frank_leveraging_2020} & 99.36 & 78.02 & 81.97 & 78.77 & 94.62 & 80.57 & 66.19 & 50.75 & 63.42 & 54.13 & 45.87 & 38.79 & 39.21 & 77.80 & 40.30 & 34.70 & 51.23 & 63.28 \\
Fusing~\cite{ju_fusing_2022} & \textbf{100.0} & 85.20 & 77.40 & 87.00 & 97.00 & 77.00 & 83.30 & 66.80 & 49.00 & 57.20 & 52.20 & 51.00 & 51.40 & 55.10 & 51.70 & 52.80 & 55.60 & 67.63 \\
LNP~\cite{liu_detecting_2022} & 99.67 & 91.75 & 77.75 & 84.10 & 99.92 & 75.39 & 94.64 & 70.85 & 84.73 & 80.52 & 65.55 & 85.55 & 85.67 & 74.46 & 82.06 & 88.75 & 87.75 & 84.07 \\
LGrad~\cite{tan_learning_2023} & 99.83 & 91.08 & 85.62 & 86.94 & 99.27 & 78.46 & 85.32 & 55.70 & 67.15 & 66.11 & 65.35 & 63.02 & 63.67 & 72.99 & 59.55 & 65.45 & 71.30 & 75.11 \\
UnivFD~\cite{ojha_towards_2023} & 99.81 & 84.93 & \underline{95.08} & \underline{98.33} & 95.75 & \textbf{99.47} & 74.96 & 86.90 & 66.87 & 62.46 & 56.13 & 63.66 & 63.49 & 85.31 & 70.93 & 50.75 & 50.73 & 76.80 \\
DIRE-G~\cite{wang_dire_2023} & 95.19 & 83.03 & 70.12 & 74.19 & 95.47 & 67.79 & 75.31 & 58.05 & 75.78 & 71.75 & 58.01 & 49.74 & 49.83 & 53.68 & 54.46 & 66.48 & 55.35 & 67.90 \\
DIRE-D~\cite{wang_dire_2023} & 52.75 & 51.31 & 49.70 & 49.58 & 46.72 & 51.23 & 51.72 & 53.30 & \textbf{98.25} & 92.42 & \underline{89.45} & 91.24 & 91.63 & 91.90 & 90.90 & 92.45 & 91.28 & 72.70 \\
NPR~\cite{tan_rethinking_2024} & 99.79 & 97.70 & 84.35 & 96.10 & 99.35 & 82.50 & \textbf{98.38} & 65.80 & 69.69 & 78.36 & 77.85 & 78.63 & 78.89 & 78.13 & 76.11 & 64.90 & 94.10 & 83.57 \\
\midrule
PC~\cite{zhong_patchcraft_2024} & \textbf{100.0} & 92.77 & \textbf{95.80} & 70.17 & \underline{99.97} & 71.58 & 89.55 & 85.80 & 82.17 & 83.79 & \textbf{90.12} & \textbf{95.38} & \textbf{95.30} & 88.91 & 91.07 & \underline{96.60} & \underline{98.43} & 89.85 \\
\rowcolor{gray!15} PC + FAIR & 99.71 & 91.41 & 92.78 & 79.07 & 99.60 & \underline{84.34} & 93.93 & 86.05 & 91.04 & 90.17 & 88.93 & \underline{94.86} & \underline{94.80} & 88.08 & \underline{92.58} & \textbf{97.10} & \textbf{98.60} & 91.94 \\
\midrule
AIDE~\cite{yan_sanity_2025} & \underline{99.99} & \textbf{99.64} & 83.95 & \textbf{98.48} & 99.91 & 73.25 & \underline{98.00} & \underline{94.20} & 93.43 & \textbf{95.09} & 77.20 & 93.00 & 92.85 & \textbf{95.16} & \textbf{93.55} & \underline{96.60} & 97.05 & \textbf{93.02} \\
\rowcolor{gray!15} AIDE + FAIR & \underline{99.99} & \underline{99.43} & 83.50 & 97.84 & \textbf{100.0} & 75.29 & 97.33 & \textbf{95.55} & \underline{93.53} & \underline{93.73} & 72.61 & 90.53 & 90.42 & \underline{94.54} & 91.81 & 95.20 & 95.13 & \underline{92.14} \\
\bottomrule
\end{tabular}
}
\end{table}

\end{document}